# SFD-Mamba2Net: Strcture-Guided Frequency-Enhanced Dual-Stream Mamba2 Network for Coronary Artery Segmentation

Nan Mu[a,b,#], Ruiqi Song[a,#], Zhihui Xu[c], Jingfeng Jiang[d], and Chen Zhao[e,*]

[a]College of Computer Science, Sichuan Normal University, Chengdu, Sichuan 610101, China.
[b]Education Big Data Collaborative Innovation Center of Sichuan 2011, Chengdu, Sichuan 610101, China.
[c]Department of Cardiology, The First Affiliated Hospital with Nanjing Medical University, Nanjing, China
[d]Department of Biomedical Engineering, Michigan Technological University, Houghton, MI 49931, USA.
[e]Department of Computer Science, College of Computing and Software Engineering, Kennesaw State University, Marietta, GA 30060, USA.

## Abstract

**Background:** Coronary Artery Disease (CAD), caused by arterial stenosis or occlusion leading to myocardial ischemia, is one of the leading causes of death worldwide. Invasive Coronary Angiography (ICA), regarded as the gold standard for CAD diagnosis, necessitates precise vessel segmentation and stenosis detection. However, ICA images are typically characterized by low contrast, high noise levels, and complex, fine-grained vascular structures, which pose significant challenges to the clinical adoption of existing segmentation and detection methods.
**Objective:** This study aims to improve the accuracy of coronary artery segmentation and stenosis detection in ICA images by integrating multi-scale structural priors, state-space-based long-range dependency modeling, and frequency-domain detail enhancement strategies.
**Methods:** We propose SFD-Mamba2Net, an end-to-end framework tailored for ICA-based vascular segmentation and stenosis detection. In the encoder, a Curvature-Aware Structural Enhancement (CASE) module is embedded to leverage multi-scale responses for highlighting slender tubular vascular structures, suppressing background interference, and directing attention toward vascular regions. In the bottleneck module, we design an Axial-Alternating Dual-Stream Mamba2 (AA-DS Mamba2) module, which incorporates horizontal–vertical dual pathways and bidirectional Mamba2 parallel sequence modeling. Within the State-Space Duality (SSD) framework, this module efficiently captures long-range spatial dependencies, thereby enhancing the representation of complex vascular topologies and fine branches. In the decoder, we introduce a Progressive High-Frequency Perception (PHFP) module that employs multi-level wavelet decomposition and deep convolution to progressively refine high-frequency details while integrating low-frequency global structures. This enables multi-scale reconstruction of vascular trunks and small branches, ultimately improving boundary detail recovery and structural connectivity.

---

[#]N. Mu and R. Song contributed equally to this work.

[*]Corresponding author to provide e-mail: czhao4@kennesaw.edu (C. Zhao).

**Results and Conclusions:** On a large-scale, multi-view ICA dataset, SFD-Mamba2Net consistently outperformed state-of-the-art methods across eight segmentation metrics, and achieved the highest true positive rate (TPR = 0.60) and positive predictive value (PPV = 0.64) in stenosis detection. These findings demonstrate that SFD-Mamba2Net provides superior performance in coronary artery segmentation and stenosis detection from ICA images, underscoring its strong potential for clinical application.

1. **Introduction**

Coronary Artery Disease (CAD), caused by the accumulation of atherosclerotic plaques that narrow or block the vascular lumen, results in insufficient myocardial blood supply and represents the leading cause of cardiovascular-related mortality worldwide [1]. It is estimated that CAD is responsible for more than nine million deaths annually, posing a serious threat to public health. Invasive Coronary Angiography (ICA), recognized as the gold standard for CAD diagnosis and therapeutic decision-making [2], enables direct assessment of coronary anatomy and the degree of stenosis. However, conventional manual interpretation of ICA images is not only time-consuming but also highly subjective and inconsistent, being influenced by physician experience, viewing angles, and the complex morphology of vessels [3], which compromises both the stability and accuracy of stenosis evaluation.

In recent years, automated vascular segmentation and stenosis detection techniques have gained increasing attention [4, 5], as they provide objective and reproducible quantitative evidence for the diagnosis of coronary artery disease (CAD). Nevertheless, segmentation and stenosis analysis of ICA images still face three fundamental challenges. <u>First</u>, low contrast and high noise levels often blur vessel boundaries, resulting in unstable segmentation outcomes. <u>Second</u>, the complex morphology and pronounced multi-scale branching of coronary vessels，particularly for micro-vessels with diameters smaller than two pixels，pose significant difficulties for accurate delineation. <u>Third</u>, variations in viewing angles and motion artifacts caused by cardiac activity and projection overlap lead to diverse vessel appearances, thereby constraining the generalization capability of computational models.

Existing deep learning-based vascular segmentation approaches primarily rely on U-Net [6] or Transformer architectures [7, 8]. While U-Net excels at local feature extraction, its limited receptive field hampers the capture of long-range spatial dependencies, often leading to vessel discontinuities or artifacts at branching points (see the red rectangle in Figure 1(c) of TransUnet [9]). In contrast, Transformers leverage self-attention mechanisms to model global relationships and effectively capture distant dependencies; however, their computational complexity and memory consumption scale quadratically with image resolution, limiting their applicability to high-resolution ICA images. Recently, structured State Space Model (SSM) [10], e.g., Mamba networks [11, 12], have demonstrated efficient global modeling capabilities for long-sequence tasks. Nonetheless, most existing SSM designs are primarily developed for Natural Language Processing (NLP) or other 1-D sequential data. When directly applied to 2-D medical images (e.g., VM-UNet [13]), they tend to overlook spatial dependencies and fail to capture the intrinsic topology of coronary vessels (see the yellow arrow in Figure 1(d)). Consequently, effectively extending the Mamba architecture to ICA image segmentation remains an open and challenging research problem.

Beyond spatial-domain approaches, frequency-domain processing has demonstrated distinct advantages in vascular segmentation [14, 15]. By leveraging frequency-domain operations such as Fourier transforms, vascular structural features can be analyzed across multiple scales, emphasizing vessel edges and branch details while suppressing background interference, thereby improving segmentation stability and contrast. Nevertheless, existing frequency-domain methods exhibit notable limitations: most rely on fixed filters(e.g., FreMAE [16]), lacking adaptive modeling of frequency components and cross-level information integration, which hinders the simultaneous capture of global vascular topology and fine branch characteristics (see the green arrow in Figure 1(e)). Moreover, due to the inherently elongated tubular morphology of vessels and their susceptibility to noise, models without effective structural priors struggle to accurately capture small vessels [17], particularly in low-contrast, high-resolution ICA images, where this challenge is most pronounced.

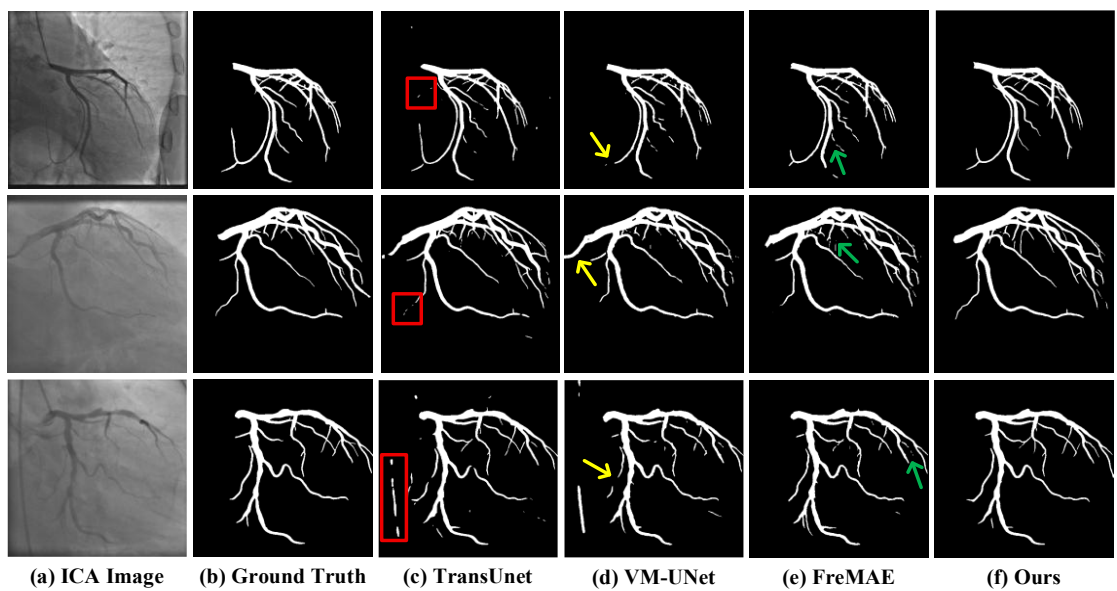

(a) ICA Image  (b) Ground Truth  (c) TransUnet  (d) VM-UNet  (e) FreMAE  (f) Ours

Figure 1. Visualization of representative failure cases from other segmentation models. (a) Original ICA image; (b) Corresponding ground-truth segmentation. (c–e) Results obtained by representative models: an attention-based U-Net variant (TransUnet [9]), an SSM-based model (VM-UNet [13]), and a frequency-domain model (FreMAE [16]). (f)

Result produced by the proposed SFD-Mamba2Net.

To address the multifaceted challenges in ICA vessel segmentation and stenosis detection, we propose SFD-Mamba2Net (Structure-Guided Frequency-Enhanced Dual-Stream Mamba2 Network), an end-to-end framework that integrates structural priors, dual-stream state-space modeling, and frequency-domain feature enhancement to achieve precise vessel segmentation and accurate stenosis assessment. In the shallow layers, the network incorporates a Curvature-Aware Structural Enhancement (CASE) module, which provides geometric priors of vessels. By leveraging multi-scale responses, this module accentuates regions with local curvature variations, guiding the model to focus on elongated tubular vessels, enhancing contrast while suppressing background noise. At the bottleneck, we introduce an Axial-Alternating Dual-Stream Mamba2 (AA-DS Mamba2) module, which employs horizontal and vertical dual pathways along with forward and backward parallel Mamba2 sequences. Under the structured State Space Duality (SSD) framework, this design efficiently captures long-range spatial dependencies, enhancing the global representation of complex vascular topologies and fine branches. Furthermore, in the decoding stage, a Progressive High-Frequency Perception (PHFP) module is employed. Through multi-level wavelet decomposition and deep convolution, this module progressively enhances high-frequency details while integrating low-frequency global structures, enabling multi-scale reconstruction of coronary vessels and substantially improving segmentation accuracy, fine vessel edge recovery, and the robustness of stenosis detection. Our code is publicly available at https://github.com/chenzhao2023/SFDMamba2ICA_seg.

In summary, the main contributions of this work are as follows：

1) We propose SFD-Mamba2Net, an end-to-end framework for ICA vessel segmentation and stenosis detection that integrates structural prior guidance, dual-stream state-space modeling, and frequency-domain feature enhancement. Systematic evaluation on a large-scale, multi-view ICA dataset demonstrates that SFD-Mamba2Net significantly outperforms existing methods in segmentation and stenosis detection, highlighting its strong potential for clinical application.

2) We design the CASE module, which captures elongated tubular vessel structures through multi-scale responses, providing structural priors that guide the network to focus on vascular regions while suppressing background interference.

3) We develop the AA-DS Mamba2 module, which employs a dual-path parallel design along with forward and backward state-space modeling. This enables efficient capture of long-range spatial dependencies, enhances the representation of complex vascular topologies, and mitigates fragmentation in vessel bifurcations and fine branches.

4) We introduce the PHFP module, which progressively enhances high-frequency sub-bands through multi-level wavelet decomposition and deep convolution while preserving low-frequency global structures. This allows multi-scale reconstruction of major vessels and fine branches, effectively refining boundary details and maintaining overall vascular connectivity.

## 2. Related Work

Current medical image segmentation approaches are predominantly built upon convolutional neural networks (CNNs) [18], with U-Net and its numerous variants emerging as the dominant paradigms. While U-Net has demonstrated remarkable effectiveness in extracting local features, its inherently limited receptive field restricts its capacity to capture long-range contextual dependencies. To overcome this limitation, recent research has increasingly explored alternative strategies, including attention mechanisms, state space models, and frequency-domain feature enhancement, with the goal of advancing both segmentation accuracy and structural detail preservation. In what follows, we present a comprehensive survey of these developments from the above three mentioned perspectives.

*2.1. Attention-based Medical Image Segmentation*

Attention mechanisms explicitly model feature correlations, thereby strengthening the network's capacity to focus on salient regions and capture long-range dependencies. These mechanisms have been widely adopted to enhance U-Net and its variants, particularly in terms of fine-grained localization and structural preservation. Building on this principle, a variety of attention-enhanced segmentation frameworks have been proposed. For example, Chen *et al*. [9] introduced TransUNet, which incorporates Transformer layers to effectively model long-range contextual dependencies. Similarly, Wang *et al*. [19] presented MT-UNet, where hybrid Transformer modules are employed to improve segmentation performance in structurally complex regions. In addition, Hu *et al*. [20] developed Perspective+UNet, which combines dual-path convolutional fusion with non-local attention to refine vascular boundary delineation. More recently, Lei *et al*. [21] proposed ConDSeg, which integrates contrast-driven attention and semantic decoupling strategies to further boost segmentation accuracy.

Although attention-based models exhibit notable strengths in capturing long-range dependencies, they continue to encounter critical limitations in medical image segmentation. Specifically, their ability to represent fine-grained local structures is often inadequate, hindering the accurate modeling of complex anatomical relationships. Moreover, in ICA images with low contrast, intricate morphology, or subtle pathological variations, these approaches are prone to structural information loss and mis-segmentation. In addition, the significant computational and memory costs associated with processing long sequences impose further challenges, thereby restricting the scalability and practicality of such methods in clinical settings.

*2.2. State Space Model-Based Medical Image Segmentation*

The State Space Model (SSM) formalizes sequence modeling as a combination of state transitions and observation mappings, thereby enabling the efficient capture of long-range dependencies. Recently, SSMs have received growing interest in their linear-time complexity and strong capability in long-sequence modeling, and they have been introduced into medical image segmentation to mitigate the deficiencies of convolution and attention mechanisms in modeling distant spatial relations. Representative advances include S4 by Gu *et al*. [10], which leverages implicit function theory to enhance long-sequence efficiency, and Mamba by Dao *et al*. [12], which markedly improves both vision and sequence tasks through selective state-space modeling and a selective scanning strategy. Building on these developments, Ruan *et al*. [13] incorporated a Visual State Space (VSS) module into VM-UNet, thereby strengthening U-Net's modeling of long-range dependencies; more recently, Wang *et al*. [22] proposed Mamba-UNet, a fully SSM-based architecture grounded in Visual Mamba that employs skip connections to enrich multi-scale detail representation, yielding further gains in segmentation accuracy.

However, existing SSMs are primarily designed for natural image processing or general computer vision tasks and lack tailored adaptations for the multi-scale structural features and complex morphologies inherent in medical images. When applied to high-resolution vascular images that are sensitive to structural details and exhibit long-range spatial dependencies, conventional SSMs tend to overlook fine branches and complex topological structures, thereby compromising the continuity and integrity of vessel segmentation.

*2.3. Frequency-Based Medical Image Segmentation*

Frequency-domain approaches transform images from the spatial domain to the frequency domain, enabling the modeling and enhancement of different frequency components, thereby emphasizing target edges, suppressing noise, and preserving structural details. In recent years, these methods have been increasingly applied to medical image segmentation, providing a powerful complement to conventional spatial-domain features. Specifically, Kong *et al*. [23] proposed DFFN, which employs a frequency-domain feature feedforward network to effectively capture and exploit the frequency information of images; Mu *et al*. [24] introduced FACU-Net, which utilizes frequency-domain multi-focal attention to precisely segment aneurysms and their associated arteries; Azad *et al.* [25] proposed FRCU-Net, incorporating Discrete Cosine Transform (DCT) features in the decoding stage to reinforce the frequency information of vessel edges; Chen *et al*. [26] developed FDConv, which dynamically learns convolutional kernel frequency responses in the Fourier domain to achieve diverse and efficient convolution operations; and Nam *et al*. [27] introduced MADGNet, integrating multi-frequency and multi-scale attention mechanisms to enhance segmentation generalizability.

Although frequency-domain representations have shown significant promise in improving contrast, suppressing noise, and maintaining fine structural details, most existing methods predominantly depend on fixed transformations and fail to dynamically adapt to multi-scale and multi-frequency components. Consequently, these approaches struggle to effectively handle the complex vessel morphologies and blurred boundaries frequently encountered in ICA images.

## 3. SFD-Mamba2Net

The overall workflow of the proposed SFD-Mamba2Net is illustrated in Figure 2. Given an input ICA image, the network first employs MASE to extract multi-scale vascular structure priors, thereby providing explicit guidance for subsequent feature modeling. A five-stage encoder then progressively compresses the spatial dimensions while capturing hierarchical semantic representations. At the bottleneck, AA-DS Mamba2 leverages dual-axial dual-stream modeling with cross-path feature interactions to comprehensively capture both local details and global dependencies and efficiently propagate them to the decoder to enhance vascular structure reconstruction. During decoding, PHFP is introduced at the beginning of each upsampling stage to fuse the high- and low-frequency components of the encoder features, which effectively restores vessel boundaries and fine-grained structures, ultimately yielding an accurate coronary artery segmentation map. Finally, the resulting segmentation is passed into the stenosis detection module, which outputs a comparative analysis highlighting stenotic regions.

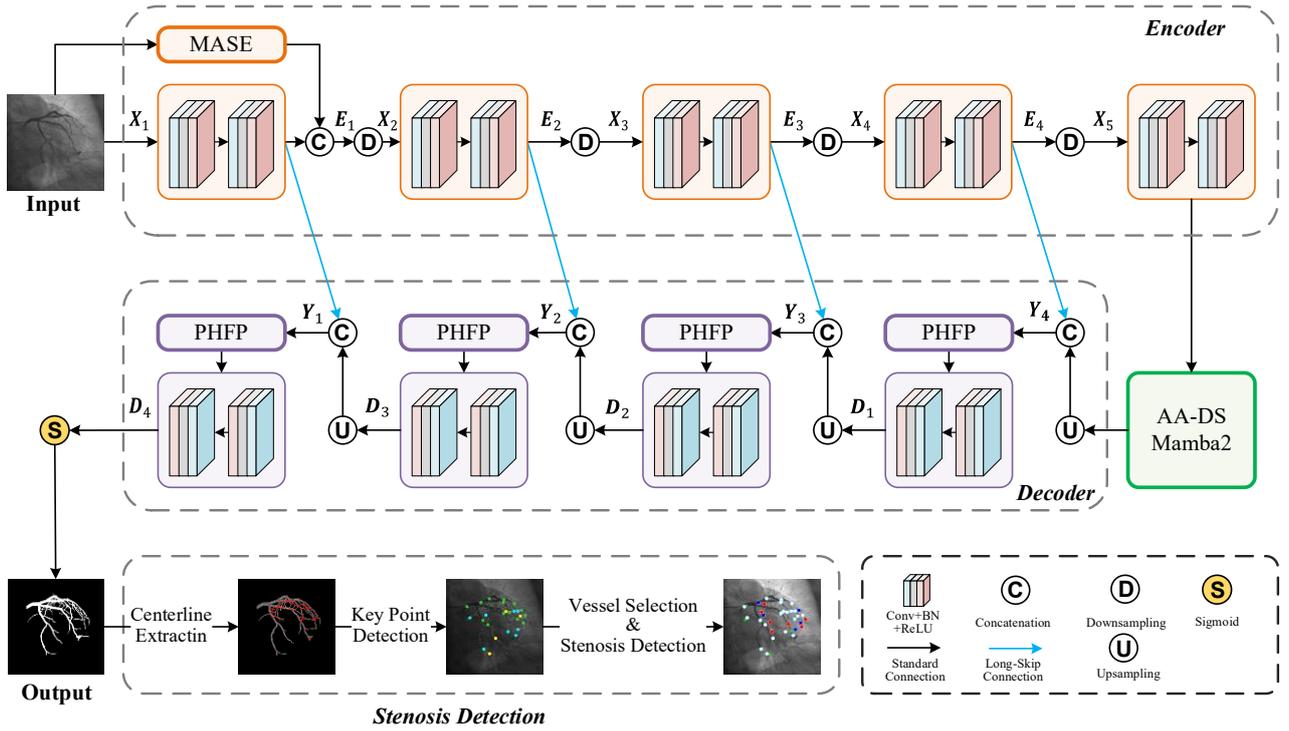

Figure 2. Overview of the proposed SFD-Mamba2Net framework. The architecture adopts an asymmetric encoder-decoder design, integrating the MASE, AA-DS Mamba2, and PHFP modules to jointly strengthen vascular structure representation, spatial dependency modeling, and boundary reconstruction, ultimately improving segmentation accuracy and enhancing the robustness of stenosis detection.

## 3.1. Encoder-Decoder Structure

As illustrated in Figure 2, the proposed SFD-Mamba2Net employs an asymmetric encoder–decoder architecture to achieve multi-scale vascular feature extraction and fine-grained structural reconstruction. The encoder is composed of four progressively stacked encoder blocks, each consisting of two CBL units (Convolution, Batch Normalization, and LeakyReLU). Specifically, each CBL unit applies a $3 \times 3$ convolution, followed sequentially by batch normalization [28] and a LeakyReLU [29] activation. To enhance regularization, a dropout layer [30] with a rate of 0.3 is additionally inserted after each batch normalization.

Notably, the output of the first encoder block is concatenated with the output of the MASE module along the channel dimension before being downsampled, whereas the subsequent encoder blocks directly perform a single downsampling operation. This hierarchical design facilitates the progressive extraction of higher-level semantic representations while simultaneously reducing spatial resolution, thereby enabling effective multi-scale feature learning. For the $i$-th encoder layer, the input feature map is denoted as $X_i \in \mathbb{R}^{C_i \times H_i \times W_i}$, where $i \in \{1,2,3,4,5\}$, and $C_i$, $H_i$ and $W_i$ represent the number of channels, height, and width, respectively. After passing through two consecutive CBL units, the output can be expressed as shown in Eq. 1.

$$E_i = \text{LeakyReLU}(\text{BN}(\text{Conv}_{3\times 3}(\text{LeakyReLU}(\text{BN}(\text{Conv}_{3\times 3}(X_i))))). \qquad (1)$$

In the encoder, downsampling is performed using a $3 \times 3$ convolution with a stride of 2, which reduces the spatial resolution by half while preserving the number of channels. The operation is formally defined in Eq. 2.

$$X_{i+1} = \text{LeakyReLU}(\text{BN}(\text{Conv}_{3\times 3}^{\text{stride}=2}(E_i))). \qquad (2)$$

By progressively compressing the spatial dimensions, the encoder is able to extract increasingly abstract and semantically enriched features, thereby providing reliable representations for the decoder.

At the network bottleneck (i.e., 5-th encoder layer), we introduce the AA-DS Mamba2 module to strengthen the modeling of complex spatial structures. Given that the bottleneck layer exhibits a relatively low spatial resolution but strong channel expressiveness, it is well suited for sequence modeling. The AA-DS Mamba2 adopts a dual-axial asymmetric design, unfolding the features along both the horizontal and vertical directions, thereby capturing long-range dependencies through sequential modeling. This module substantially enhances the model's global perception of target structures (e.g., vessels and tubular tissues), thereby improving the completeness and boundary consistency of vascular segmentation in ICA images.

The decoder is composed of four sequential decoder blocks. At the $i$-th layer, the input feature map $Y_i \in \mathbb{R}^{C_i \times H_i \times W_i}$, $i \in \{1,2,3,4\}$, is obtained by concatenating the feature $E_i$ from the corresponding encoder layer with the upsampled feature $U_{i+1}$ from the preceding decoder layer along the channel dimension, as shown in Eq. 3.

$$Y_i = Concat(U_{i+1}, E_i), \qquad (3)$$

where the upsampled feature $U_{i+1}$ is generated by applying nearest-neighbor interpolation to the output of the previous decoder layer $D_{i+1}$, followed by a $1 \times 1$ convolution to reduce the channel dimension, as denoted in Eq. 4.

$$U_{i+1} = \text{Conv}_{1\times 1}\bigl(\text{Interpolate}(D_{i+1})\bigr). \qquad (4)$$

To further enhance the representation of global contextual and structural information, each decoder stage begins with the incorporation of a PHFP module, which emphasizes high-frequency components and facilitates the precise recovery of vascular boundaries and fine-grained structures. The frequency-enhanced features are subsequently passed through the CBL unit within each decoder block, thereby enabling deeper semantic extraction and improving nonlinear modeling capability.

Overall, in the context of coronary artery segmentation, this architectural design enables more effective restoration of spatial details and reinforcement of boundary information, which in turn substantially improves segmentation accuracy and preserves vascular structural integrity.

*3.2. Curvature-Aware Structure Enhancer (CASE) Module*

To enhance the perception of slender vascular structures in invasive coronary angiography (ICA) images, we introduce a Curvature-Aware Structure Enhancer (CASE) module at the input stage of the SFD-Mamba2Net backbone. Specifically, the CASE module leverages multi-scale Hessian filtering to incorporate structural priors, thereby strengthening the representation of regions with local curvature variations and improving the model's sensitivity to linear structures. By computing the eigenvalues of the Hessian matrix across multiple scales, the CASE module effectively captures subtle local geometric curvatures, which in turn highlights regions with pronounced linear morphology (e.g., blood vessels). This mechanism not only amplifies the salience of elongated targets but also suppresses background noise and irrelevant non-structural regions. Moreover, unlike conventional deep networks that rely purely on data-driven learning, the CASE module introduces explicit structural priors with clear physical significance at an early stage of the network, providing interpretable geometric cues that guide subsequent feature extraction toward more precise discrimination between vessels and background.

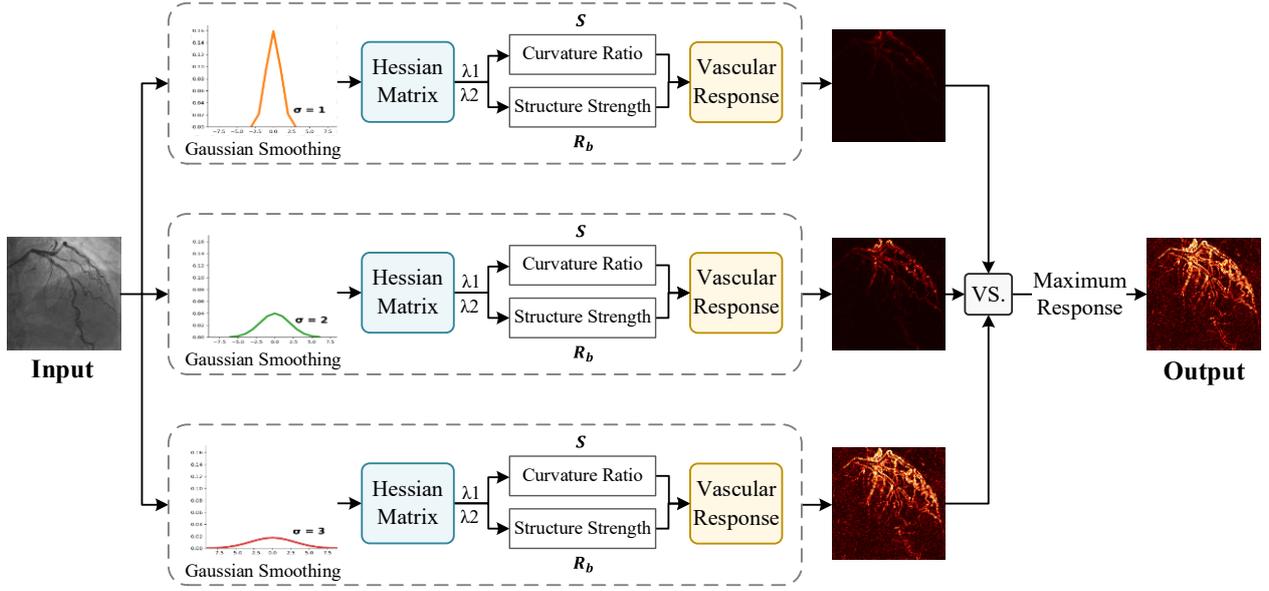

Figure 3. Schematic Diagram of the CASE Module.

As illustrated in Figure 3, the CASE module first applies three Gaussian kernels with standard deviations σ ∈ {1, 2, 3}, $G(x,y) = \frac{1}{2\pi\sigma^2}\exp\left(-\frac{x^2+y^2}{2\sigma^2}\right)$ to smooth the input image, thereby reducing noise and enhancing structural responses. Considering the significant variation in vessel radius in ICA images, multi-scale Gaussian smoothing is employed to adapt to vascular structures of different spatial scales. Subsequently, the smoothed image is convolved with the Hessian operator to obtain the approximate Hessian matrix at each pixel is denoted in Eq. 5.

$$H(x,y) = \begin{bmatrix} D_{xx} & D_{xy} \\ D_{yx} & D_{yy} \end{bmatrix}, \tag{5}$$

where the second-order derivatives $D_{xx}$, $D_{xy}$, $D_{yx}$, and $D_{yy}$ are computed using Sobel differential kernels. The two eigenvalues λ₁ and λ₂ are then analytically calculated in Eq. 6.

$$\lambda_{1,2} = \frac{D_{xx}+D_{yy}}{2} \pm \sqrt{\frac{(D_{xx}-D_{yy})^2}{4} + D_{xy}^2} + \varepsilon, \tag{6}$$

where $|\lambda_1| \geq |\lambda_2|$ are sorted by absolute value, characterizing the principal and secondary curvatures of the structure. The vascular response is subsequently computed using the vesselness function defined in Eq. 7.

$$Vesselness(x) = \begin{cases} 0 & if\ \lambda_2 > 0 \\ \exp\left(-\frac{R_b^2}{2\beta^2}\right) \cdot \left(1 - \exp\left(-\frac{S^2}{2c^2}\right)\right) & otherwise \end{cases}, \tag{7}$$

where the curvature ratio $R_b = \frac{|\lambda_1|}{|\lambda_2|+\epsilon}$ quantifies the elongation of the structure, and the structure strength $S = \sqrt{\lambda_1{}^2 + \lambda_2{}^2}$ measures its prominence; $\beta$ and $c$ are hyperparameters controlling sensitivity. When $\lambda_2 > 0$, typically corresponding to bright background regions or non-elongated structures, the response is set to zero to effectively suppress interference from non-vascular regions. Conversely, when $\lambda_2 < 0$, indicating potential linear structures or edges, the response is jointly determined by the two exponential terms $R_b$ and $S$:

As illustrated in Figure 4, when the curvature ratio $R_b$ approaches 0, it indicates a highly linear structure corresponding to vascular regions; conversely, when $R_b$ approaches 1, the structure tends toward a circular or non-directional shape, and the response approaches zero. On the other hand, a large structure strength $S$ indicates that the region exhibits prominent structural characteristics. The multiplication of these two factors ensures that the $Vesselness(x)$ generates high responses only when the structure is both linear and salient.

In practice, the CASE module computes the vesselness response in parallel across multiple scales and selects the maximum response at each pixel as the output, thereby substantially enhancing vascular structures at different scales, including micro-vessels and capillaries. This mechanism effectively suppresses background, speckle, and texture noise, improving the clarity and continuity of structural representations.

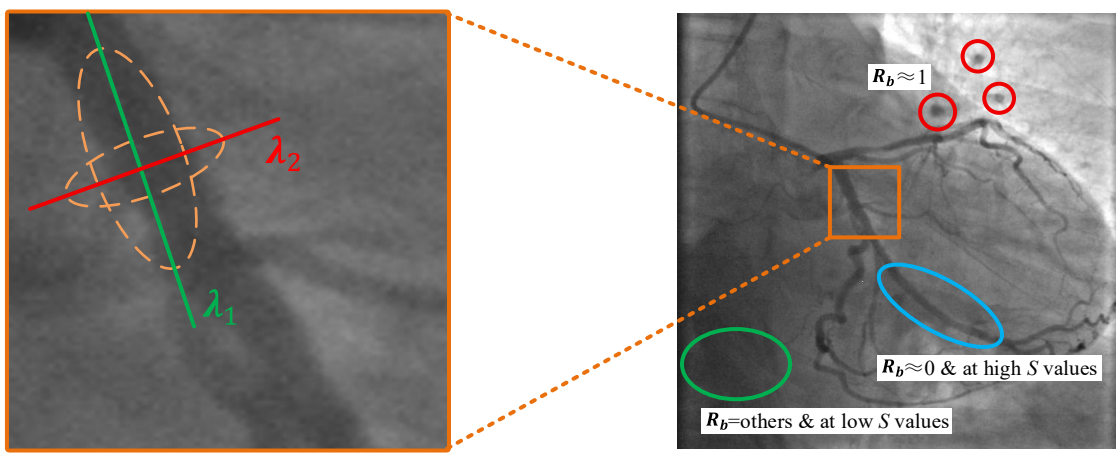

Figure 4. Visualization of the vesselness response mechanism. (a) Local zoom-in of vascular regions illustrating the correspondence between Hessian eigenvalues $(\lambda_2, \lambda_2)$ and linear structures; (b) Variation trends of $R_b$ and $S$ values across different locations in ICA images.

After the initial feature extraction in the backbone network, we incorporated the Curvature-Aware Structure Enhancer (CASE) module to enhance the model's sensitivity to elongated structures. This module processes the raw input image in parallel across multiple scales, computing vesselness response maps at each scale to effectively reinforce coherent and slender vascular structures, while substantially suppressing background artifacts, noise, and non-structural regions. The resulting multi-scale structural prior maps are concatenated along the channel dimension with the low-level features extracted by the first convolutional layer, thereby integrating local geometric priors with early visual representations and providing explicit guidance for structural information.

To further improve the discriminability of the fused features, we introduce a lightweight channel attention mechanism that adaptively weights the multi-channel features, allowing the model to focus on structurally salient regions and enhancing the structural sensitivity of the feature representation. Notably, the CASE module is based on multi-scale Hessian filtering, requires no learnable parameters, and offers efficient and stable responses, enabling the extraction of elongated, coherent structures in images with uneven intensities without imposing additional learning burden.

Considering that coronary arteries in ICA images generally exhibit thin, continuous linear morphology, the multi-scale vesselness responses extracted by the CASE module comprehensively cover vessels of varying widths and select the optimal scale for response enhancement. This effectively highlights "vessel-like" regions while suppressing speckle noise and non-structural background interference. Ultimately, the module substantially improves the model's sensitivity to linear structures, providing a robust structural representation foundation for subsequent segmentation tasks.

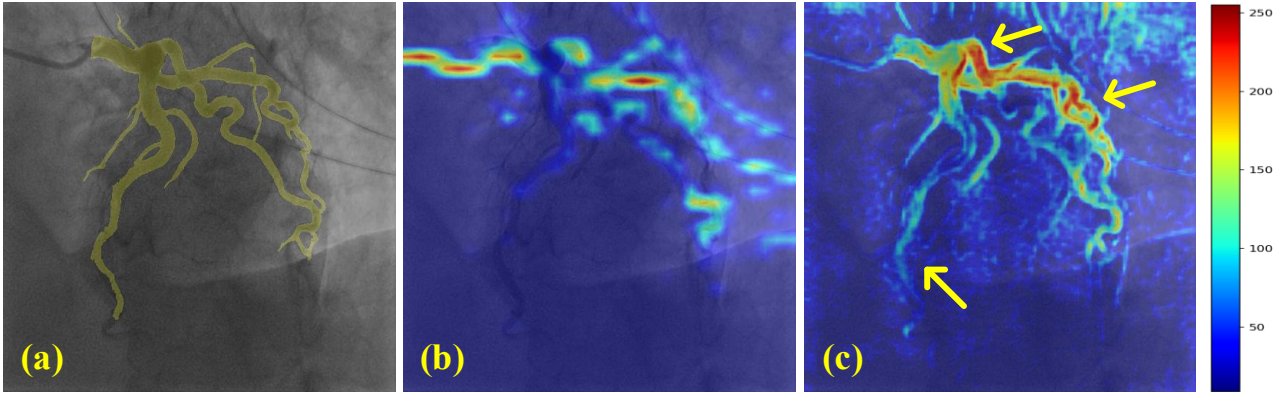

Figure 5. Visualization demonstrating the effectiveness of the CASE module. (a) Overlay of the input image and ground-truth labels; (b) and (c) Grad-CAM heatmaps [31] generated by standard convolution and the CASE module, respectively. Red regions indicate strong activations most critical to the decision, yellow and green indicate moderate importance, and blue denotes minimal contribution.

It is worth emphasizing that, since the CASE module constructs explicit structural priors based on geometric models, it is capable of enhancing linear structures even in regions with low contrast and poor signal-to-noise ratios. This characteristic is particularly critical for mitigating common issues in ICA images, such as disrupted vascular topology and background interference. As illustrated in the heatmaps of Figure 5, the practical effectiveness of this approach is evident: in Figure 5(c), the vascular structures indicated by the yellow arrows are markedly enhanced, exhibiting clearer and more continuous contours, which demonstrates the CASE module's ability to reinforce structural information under small-scale and low-contrast conditions.

Overall, the CASE module not only effectively suppresses non-structural background regions and noise interference but also significantly amplifies responses in vascular regions, enabling the model to acquire strong structural awareness at shallow stages. This early-stage structural enhancement guides subsequent deep feature extraction to focus more precisely on key anatomical regions, thereby establishing a solid foundation for improved segmentation accuracy and discriminative capability.

*3.3. Axial-Alternating Dual-Stream Mamba2 (AA-DS Mamba2)*

To enhance the model's spatial modeling capabilities in ICA image segmentation tasks, we propose a novel state-space-based architecture termed Axial-Alternating Dual-Stream Mamba2

(AA-DS Mamba2). This module employs a dual-axial asymmetric design, consisting of two complementary paths that independently model long-range dependencies along the horizontal (width) and vertical (height) directions. Within each path, two parallel Mamba2 streams process both the original sequence and its flipped counterpart, thereby capturing directionally symmetric contextual information under the State Space Duality (SSD) framework [11]. This design not only preserves the computational efficiency of Mamba2 in long-sequence modeling, but also substantially enhances the model's adaptability to complex spatial structures, making it particularly suitable for ICA vascular segmentation tasks characterized by intricate details and long-range dependency features.

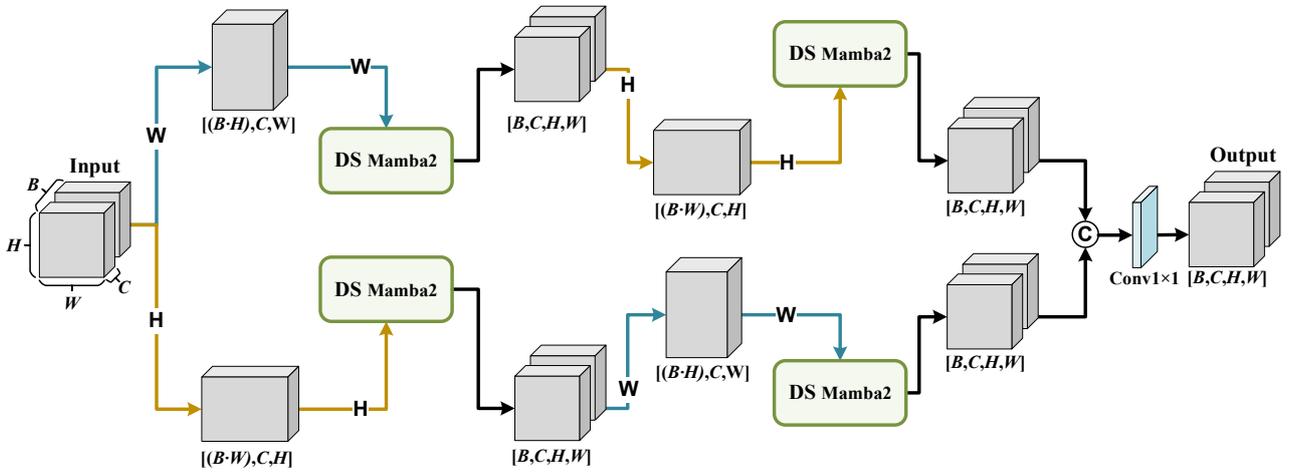

Figure 6. Schematic illustration of the proposed AA-DS Mamba2 architecture.

As shown in Figure 6, the AA-DS Mamba2 module consists of two spatially complementary axial paths, which sequentially model features along the "width→height" and "height→width" directions. This alternating axial design enables the network to fully capture structural heterogeneity and directional context in both horizontal and vertical dimensions. Within each path, two DS Mamba2 submodules are cascaded, and each submodule processes both the original and flipped sequences in parallel, thereby constructing bidirectional dependencies along the temporal dimension in the state space and strengthening global context modeling. Finally, the outputs of the dual paths are concatenated along the channel dimension and fused through a $1 \times 1$ convolution, producing feature maps that integrate multi-directional semantics and spatial context, effectively enhancing segmentation accuracy in fine structures such as tiny arteries.

For the input feature map $X \in \mathbb{R}^{B \times C \times H \times W}$, where $B$ denotes the batch size, $C$ is the number of channels, $H$ is the height, and $W$ is the width, the AA-DS Mamba2 module first flattens it along the width (horizontal axis) to form a 1-D sequence: $X_{w1} = \text{Reshape}(X, axis = W) \in \mathbb{R}^{(B \cdot H) \times C \times W}$, which is then fed into a DS Mamba2 submodule to capture long-range dependencies along the horizontal direction. The processed feature $Y_{w1}$ is reshaped back to 2-D and subsequently unfolded along the height (vertical axis): $X_{h1} = \text{Reshape}(Y_{w1}, axis = H) \in \mathbb{R}^{(B \cdot W) \times C \times H}$, and passed through another DS Mamba2 submodule to model structural information along the vertical axis. Simultaneously, to realize the axial-alternating dual-stream design, the module also unfolds the original input $X$ along the height, yielding $X_{h2} = \text{Reshape}(X, axis = H) \in \mathbb{R}^{(B \cdot W) \times C \times H}$, after processing through a DS Mamba2 submodule, the resulting output $Y_{h2}$ is further unfolded along the horizontal axis: $X_{w2} = Reshape(Y_{h2}, axis = W) \in \mathbb{R}^{(B \cdot H) \times C \times W}$ and then passed through another DS Mamba2 submodule to model long-range dependencies along the alternate axis.

Finally, the outputs from the two alternating axial paths are concatenated along the channel dimension: $Y_{cat} = \text{Concat}(Y_1, Y_2, dim = 1) \in \mathbb{R}^{B \times 2C \times H \times W}$, followed by a $1 \times 1$ convolution to fuse features, producing the final context-enhanced feature map $Y_{out} = \text{Conv}_{1 \times 1}(Y_{cat})$.

By employing the axial-alternating dual-stream sequential processing strategy, this design effectively integrates long-range dependencies in both horizontal and vertical directions, substantially enhancing the model's capacity to represent complex spatial structures and fine-grained vascular details.

*3.3.1. DS Mamba2*

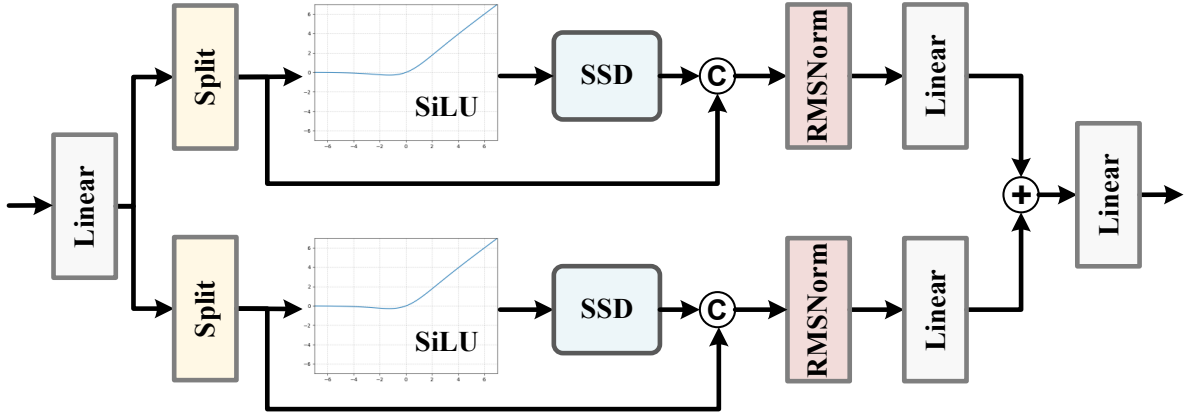

Figure 7. Schematic diagram of the proposed DS Mamba2 module.

As shown in Figure 7, the DS Mamba2 module is a dual-stream state-space architecture designed for one-dimensional sequence inputs, capable of simultaneously modeling forward and reverse state dependencies within a sequence. For an input tensor $X \in \mathbb{R}^{B \times C \times L}$, where $B$ denotes the batch size, $C$ is the number of channels, and $L$ is the sequence length. First, the input $X$ is linearly projected to the hidden dimension $d_{model}$ required by Mamba2, yielding $X_{in} = \text{Linear}_{in}(X) \in \mathbb{R}^{B \times d_{model} \times L}$. Subsequently, $X_{in}$ and its flipped sequence $Flip(X_{in})$ are respectively fed into two independent Mamba2 pathways (including Split, SiLU activation [32], SSD module, RMSNorm [33], and linear projection), to separately capture the forward and backward state dependencies. The outputs of the two streams are then fused to obtain the bidirectionally modeled result $X_{ds}$, which is subsequently projected back to the target channel dimension $C_{out}$ via a linear layer : $X_{out} = \text{Linear}_{out}(X_{ds}) \in \mathbb{R}^{B \times C_{out} \times L}$, and finally reshaped to match the original dimension order for downstream modules. This module, through bidirectional modeling of the sequence in both forward and reverse directions, significantly enhances the contextual representation of features, thereby improving the overall expressiveness and generalization capability of the network.

By employing this bidirectional modeling strategy, the DS Mamba2 significantly enhances the contextual representation of sequential features, thereby improving the overall expressiveness and generalization capability of the network.

*3.3.2. Mamba2*

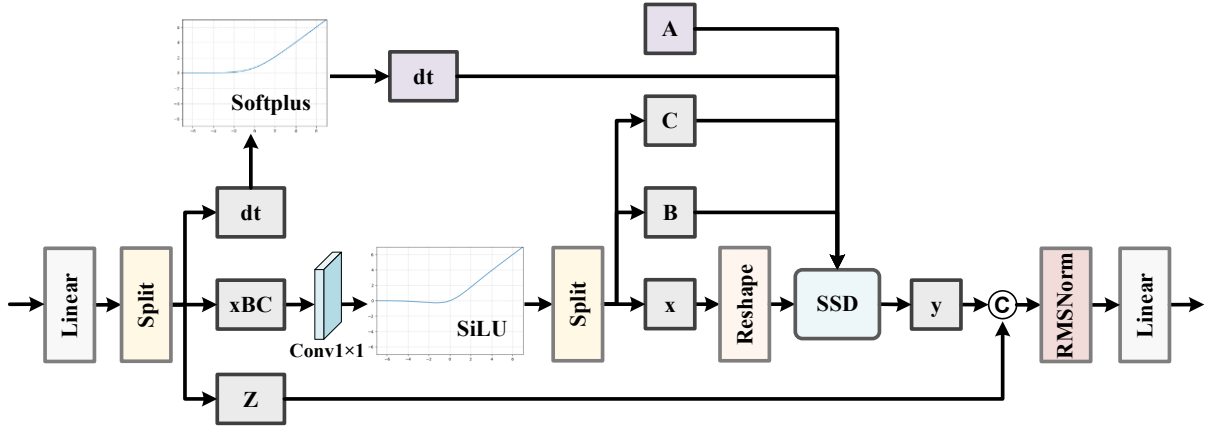

Figure 8. Structural diagram of the Mamba2 module.

Figure 8 illustrates the detailed architecture of the Mamba2 [11] module within DS Mamba2. This module operates on the input sequence by first performing feature reorganization through an input projection followed by a nonlinear activation, and subsequently executing structured state modeling under the State Space Duality (SSD) framework to progressively propagate states across the sequence and construct the output.

The inference process consists of several key stages: input feature projection, activation and splitting, SSD-based state space propagation, fusion and normalization, and final output projection. Algorithm 1 presents the complete inference workflow of Mamba2.

---

**Algorithm 1:** Mamba2

---

**Input:** u∈$\mathbb{R}^{batch\_size \times d_{model} \times sequence\_length}$

**Output:** y∈$\mathbb{R}^{batch\_size \times d_{model} \times sequence\_length}$

---

**1. InputProjection** (u)
        zxbcdt ← in_proj(u) #Apply in_proj to u to get zxbcdt
        z, xBC, dt ← Split(zxbcdt) #Split zxbcdt into z, xBC, and dt
   **Return** dt, z, xBC

**2. Activation & Splitting** (xBC, dt)
        xBC ← Conv1d(xBC.transpose(1, 2)).transpose(1, 2)[:, :u.shape[1], :]
        xBC ← SiLU(xBC)
        dt ← Softplus(dt + dt_bias)
        x, B, C ← Split(xBC) #Split the processed result into x, B, and C
        x ← Reshape(x)
   **Return** dt, x, B, C

**3. SSD (x, A, B, C, dt)** # SSD-based state space propagation
    y ← SSD(x, A, B, C, dt)
  **Return** y

**4. Fusion & Normalization (y, x, D, norm)**
    y ← Reshape(y)
    y ← y + x * D.unsqueeze(-1)
    y ← RMSNorm(y, z)
  **Return** y

**5. OutputProjection (y)**
    y ← out_proj(y)
  **Return** y

**Main Routine**
    [z, xBC, dt] ← InputProjection(u)
    [x, B, C, dt] ← Activation_And_Splitting (xBC, dt)
    y ← SSD (x, A, B, C, dt)
    y ← Fusion_And_Normalization(y, x, D, norm)
    y ← OutputProjection(y)
  **Return** y

The core design of Mamba2 is built upon the State Space Duality (SSD) theory. SSD provides a mathematical dual mechanism for efficiently transforming continuous-time state space equations into parallel structures suitable for discrete sequence modeling tasks. Specifically, SSD converts a continuous state space model of the form $\begin{cases} \dot{h}(t) = Ah(t) + Bx(t) \\ y(t) = Ch(t) \end{cases}$ into a parallel implementation optimized for discrete sequences. By explicitly modeling the state transition matrix $\bar{A}$, input matrix $\bar{B}$, and output matrix $\bar{C}$, while incorporating a learnable dynamic step size $\Delta t$, SSD captures dependencies of arbitrary length while maintaining linear time complexity. This design endows Mamba2 with a pronounced advantage in long-sequence modeling, making it particularly suitable for image tasks that require extensive spatial context. In the context of ICA vascular image segmentation, the SSD architecture significantly enhances the model's ability to capture long-range spatial dependencies, thereby improving contextual understanding and real-time inference performance when processing complex vascular topologies.

    Algorithm 2 highlights the core SSD components within the Mamba2 architecture that underpin its long-range spatial modeling capability.

| **Algorithm 2:** SSD |
|---|
| **Input:** |
|     x          # Input features, shape [B, L, H, P] (L = seq. length) |
|     A_log     # Learnable params for state transition matrix A, shape [H] |
|     B, C      # In/Out projection matrices, shapes [H, N] and [H, N] |
|     dt_bias    # Learnable bias, shape [H] |
|     chunk_size # Chunk size Q |
| 1. Preprocessing |
|     A ← -exp(A_log)                # Shape [H] |
|     dt ← softplus(dt_raw + dt_bias)    # From in_proj split, shape [B, L, H] |
|     x ← x × dt.unsqueeze(-1)        # Time-step control, shape [B, L, H, P] |
|     A ← A × dt                   # Synchronize with A update |
| 2. Chunk preparation |
|     # Split seq by chunk_size |
|     x_blocks ← reshape(x, [B, L/Q, Q, H, P]) |
|     B_blocks ← reshape(B_unsplit, [B, L/Q, Q, H, N]) |
|     C_blocks ← reshape(C_unsplit, [B, L/Q, Q, H, N]) |
|     A_blocks ← reshape(A, [B, L/Q, Q, H]) → permute to [B, H, L/Q, Q] |
| 3. In-chunk (diagonal) output computation |
|     L_diag ← Exp( SegSum(A_blocks) ) |
|     Y_diag ← einsum("bclhn, bcshn, bhcls -> bclhp", C_blocks, B_blocks, L_diag, x_blocks) |
| 4. In-chunk state computation |
|     A_cumsum ← Cumsum(A_blocks, dim=-1) |
|     decay_states ← Exp( A_cumsum[..., -1:] - A_cumsum ) |
|     states ← einsum("bclhn, bchpn, bhcl -> bchpn", B_blocks, decay_states, x_blocks) |
| 5. Cross-chunk state propagation |
|     # Pad zeros at time dim front |
|     zero_state ← zeros_like(states[..., :1]) |
|     states_padded ← concat([zero_state, states], dim=3) |
|     decay_chunk ← Exp( SegSum(pad_last(A_cumsum[..., -1])) ) |
|     states ← einsum("bhzc, bchpn -> bzhpn", decay_chunk, states_padded) |
|     states ← states[..., :-1]    # Remove padded zeros |
| 6. Cross-chunk (off-diagonal) output computation |
|     decay_out ← Exp(A_cumsum) |
|     Y_off ← einsum("bclhn, bchpn, bhcl -> bclhp", C_blocks, states, decay_out) |
| 7. Merge outputs |
|     Y_blocks ← Y_diag + Y_off |
|     Y ← reshape(Y_blocks, [B, L, H, P]) |
| **Return Y** |

It is worth emphasizing that the SSD module in Mamba2 does not adopt the conventional sequential state update strategy of traditional state space model (SSM), such as the stepwise recursion $X_{t+1} = AX_t + Bu_t$. Instead, it is restructured under the State Space Duality framework, which introduces a "block-diagonal + low-rank residual" decoupling strategy through structural reformulation of continuous-time state space systems, enabling the state evolution process to be efficiently mapped to a parallel, differentiable computational structure. This design significantly enhances both the efficiency and expressive capability of discrete sequence modeling [11].

This property is particularly crucial for vascular modeling in ICA images. Vascular structures often exhibit long-range non-local dependencies, pronounced directional characteristics, and structural discontinuities, which are challenging for conventional methods to capture effectively. In Mamba2, the SSD module addresses this by partitioning sequences into multiple chunks and separately computing intra-chunk local state outputs ($Y_{diag}$) and inter-chunk global state propagation terms ($Y_{off}$). This strategy allows the model to capture long-range spatial contextual information more effectively, thereby improving its ability to perceive complex vascular trajectories and connectivity patterns.

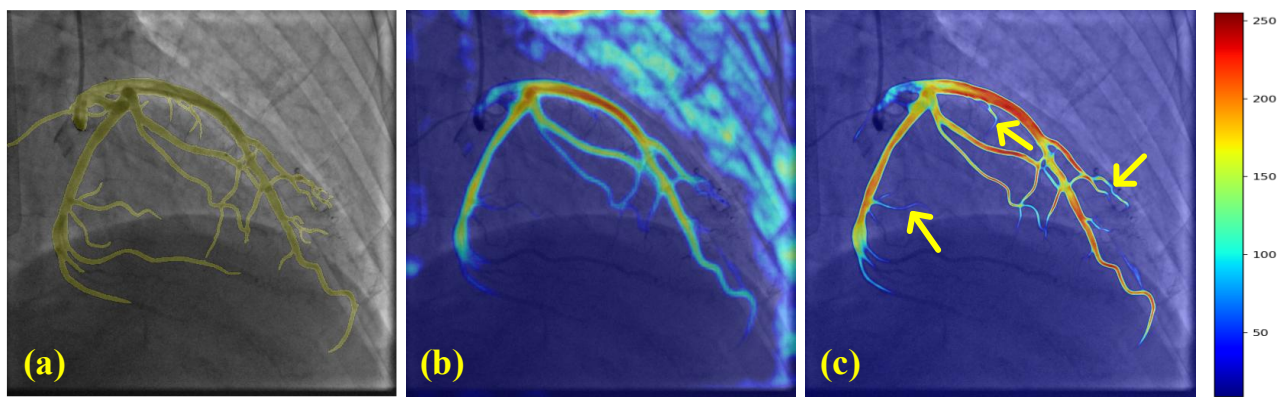

Figure 9. Visual analysis of the effectiveness of the AA-DS Mamba2 module. (a) Overlay of the input image and ground truth labels; (b) and (c) Grad-CAM heatmaps generated by Mamba2 and AA-DS Mamba2, respectively. Red regions indicate strong activations that contribute most to the model's decision, yellow and green indicate moderate activations, and blue represents areas with minimal contribution.

Although conventional Mamba2 [11] has reduced the computational complexity to linear via its parallel SSD module, its state update in high-resolution ICA images remains constrained by a fixed, one-directional scanning order. Such unidirectional modeling fails to adequately capture the structural heterogeneity of vessels across orthogonal dimensions, thereby weakening long-range spatial dependencies and causing frequent discontinuities in thin vascular segments, while simultaneously diminishing sensitivity to micro-vessels. To address these limitations, we introduce AA-DS Mamba2, which employs a cross-attentional dual-stream architecture to concurrently process "width→height" and "height→width" axial pathways, each cascaded with bidirectional DS-Mamba2 sub-modules that establish dual temporal dependencies within the state space. Crucially, this design achieves the above benefits without incurring additional memory overhead, effectively mitigating the structural deficiencies inherent in the vanilla Mamba2, and consequently delivering substantial improvements in segmenting vessels with blurred boundaries and irregular morphologies.

As illustrated in Figure 9, a comparison of Grad-CAM heatmaps generated by conventional Mamba2 (Figure 9(b)) and AA-DS Mamba2 (Figure 9(c)) clearly demonstrates the superior feature modeling capability of the latter. Figure 9(c) exhibits broader activation regions and stronger long-range dependency capture, indicating that AA-DS Mamba2 effectively models relationships between distant regions in the image. In contrast, conventional Mamba2 activations are more localized, primarily confined to specific regions, reflecting its limited capacity for long-distance structural modeling.

Moreover, the architecture of AA-DS Mamba2 not only preserves precise modeling of local sequential details but also leverages cross-chunk non-local state interactions to enhance the analysis and disentanglement of multi-scale anatomical structures. This capability enables more effective identification of vascular and tissue features across varying spatial scales. The regions indicated by the yellow arrows in Figure 9(c) further corroborate this advantage: the heatmaps exhibit richer and clearer structural details, highlighting the model's superior performance in multi-scale feature representation.

In summary, AA-DS Mamba2, with its cross-chunk continuous state propagation, efficient parallel computation framework, and outstanding multi-scale modeling capacity, serves as a critical module for addressing the combined challenges of high-resolution, detail-rich, and long-range dependency characteristics inherent in ICA image segmentation tasks. Consequently, it substantially enhances the model's performance and practical utility in complex medical imaging scenarios.

*3.4. Progressive High-Frequency Perception (PHFP) Module*

To improve the network's ability to capture fine vascular structures in ICA images, we incorporated the Progressive High-Frequency Perception (PHFP) module into the decoder of SFD-Mamba2Net. As illustrated in Figure 10, PHFP leverages a cascaded Wavelet Transform (WT) and Inverse Wavelet Transform (IWT) [34] architecture, progressively enhancing high-frequency components through deep convolutional operations. This design allows the model to more accurately perceive vascular edges, delicate branches, and stenotic lesions, while simultaneously preserving low-frequency components to maintain the overall vascular topology. Consequently, PHFP achieves a refined balance between structural integrity and fine-grained detail representation.

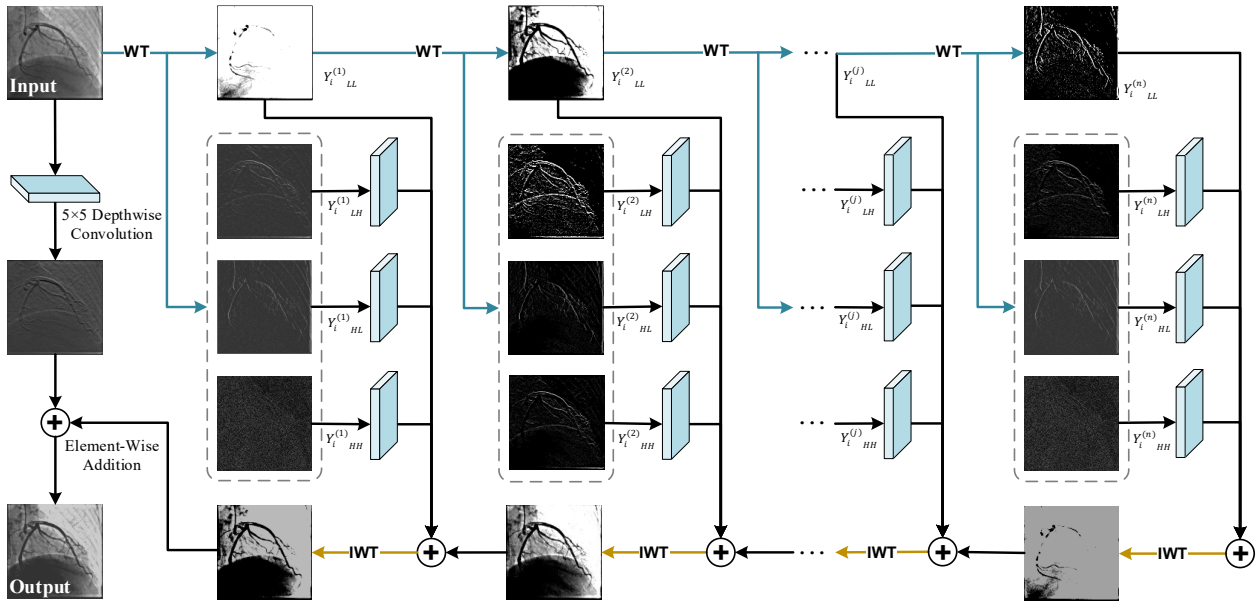

Figure 10. Schematic diagram of the PHFP module.

Specifically, the PHFP module constructs its filters based on the Haar wavelet, which consists of one low-pass filter ($f_{LL} = [1/\sqrt{2} \ \ 1/\sqrt{2}]$) and three high-pass filters ($f_{LH} = [1/\sqrt{2} \ \ -1/\sqrt{2}]$, $f_{HL} = \begin{bmatrix} 1/\sqrt{2} & 1/\sqrt{2} \\ -1/\sqrt{2} & -1/\sqrt{2} \end{bmatrix}$, $f_{HH} = \begin{bmatrix} 1/\sqrt{2} & -1/\sqrt{2} \\ -1/\sqrt{2} & 1/\sqrt{2} \end{bmatrix}$). For each decoder input feature

$Y_i$, $i \in \{1,2,3,4\}$, PHFP first applies the four filters to generate four initial sub-bands: $\{Y_{i\ LL}^{(j)}, Y_{i\ LH}^{(j)}, Y_{i\ HL}^{(j)}, Y_{i\ HH}^{(j)}\}$, $j=1$. Here, the low-frequency sub-band $Y_{i\ LL}^{(j)}$ primarily retains the global vascular structure, while the high-frequency sub-bands $Y_{i\ LH}^{(j)}, Y_{i\ HL}^{(j)}, Y_{i\ HH}^{(j)}$ capture horizontal, vertical, and diagonal texture details, respectively. To obtain finer-grained features, PHFP recursively applies the Haar wavelet transform (denoted as $WT(\cdot)$) to the low-frequency sub-band from the previous level, i.e., $Y_{i\ LL}^{(j-1)}$, $j \in \{2, \dots, n\}$, thereby progressively constructing a hierarchical high-frequency decomposition:

$$\left\{\underbrace{Y_{i\ LL}^{(j)}}_{low}, \underbrace{Y_{i\ LH}^{(j)}, Y_{i\ HL}^{(j)}, Y_{i\ HH}^{(j)}}_{high}\right\} = \begin{cases} WT(Y_i), & j = 1, \\ WT\left(Y_{i\ LL}^{(j-1)}\right), & j = 2, \dots, n \end{cases}. \quad (8)$$

At the $j$-th decomposition level, PHFP refines the high-frequency sub-bands $\{Y_{i\ LH}^{(j)}, Y_{i\ HL}^{(j)}, Y_{i\ HH}^{(j)}\}$ using a $5 \times 5$ depthwise convolution $\mathcal{D}(\cdot)$ for feature enhancement and noise suppression. These refined high-frequency components are then aggregated with the corresponding low-frequency sub-band $Y_{i\ LL}^{(j)}$ and the reconstructed feature from the next level $R_i^{(j+1)}$. The combined representation is passed through the inverse wavelet transform (denoted as $IWT(\cdot)$), yielding the reconstructed feature $R_i^{(j)}$:

$$R_i^{(j)} = \begin{cases} IWT\left(Y_{i\ LL}^{(j)} + \mathcal{D}\left(Y_{i\ LH}^{(j)}\right) + \mathcal{D}\left(Y_{i\ HL}^{(j)}\right) + \mathcal{D}\left(Y_{i\ HH}^{(j)}\right) + R_i^{(j+1)}\right), j = 1, \dots, n-1 \\ IWT\left(Y_{i\ LL}^{(j)} + \mathcal{D}\left(Y_{i\ LH}^{(j)}\right) + \mathcal{D}\left(Y_{i\ HL}^{(j)}\right) + \mathcal{D}\left(Y_{i\ HH}^{(j)}\right)\right), j = n \end{cases}. \quad (9)$$

Leveraging the linear additivity inherent to wavelet transforms, PHFP seamlessly integrates multi-level high-frequency components with their low-frequency counterparts in the frequency domain. Consequently, the representation of fine vascular details is substantially enriched while the global vascular topology remains intact. Ultimately, the reconstruction derived from the first-level map $R_i^{(1)}$ is fused with the original decoder input $Y_i$ to yield the module's final output:

$$Y_i^{PHFP} = \mathcal{D}(Y_i) + R_i^{(1)}. \quad (10)$$

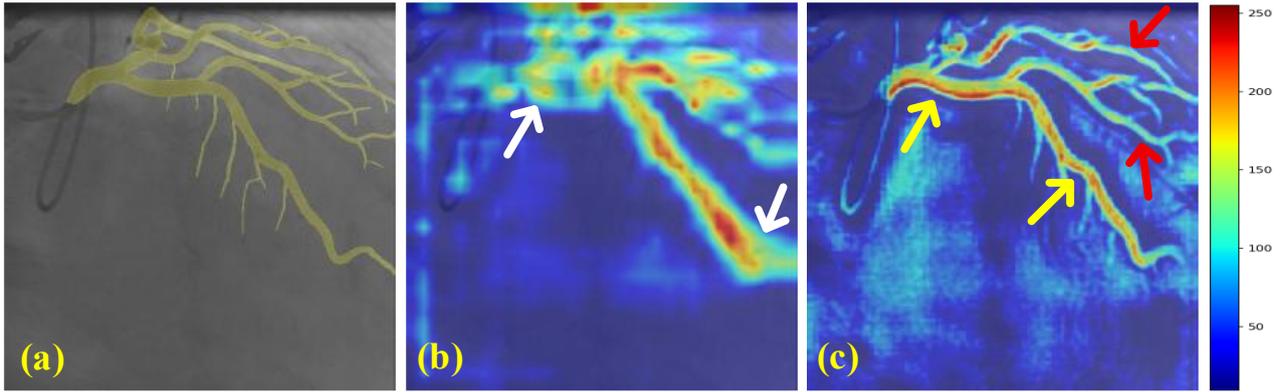

Figure 11. Visualization of the effectiveness of PHFP. (a) Overlap of the input image and ground truth; (b) and (c) Grad-CAM heatmaps generated by standard convolution and PHFP, respectively. Red regions indicate the most critical high-activation areas for model decision-making, yellow and green denote moderately important regions, and blue represents areas with minimal contribution.

Theoretically, the global topological structure of vessels in ICA images is primarily embedded within the low-frequency components, whereas fine branches and noise are predominantly distributed across the high-frequency sub-bands. The proposed PHFP module progressively enhances the high-frequency components to refine local vascular details, while simultaneously decomposing the low-frequency components to strengthen the representation of major vascular trunks and overall connectivity. This dual strategy substantially improves the robustness and accuracy of the segmentation model when confronted with complex vascular structures.

As illustrated in Figure 11, compared with the feature activation maps derived from conventional convolution, the responses generated by PHFP exhibit greater stability and concentration along the fine vascular branches, while simultaneously showing stronger activations in the main arterial trunks. This observation provides clear evidence for the effectiveness of the progressive high-low frequency decomposition and fusion strategy in vascular segmentation tasks. More specifically, by progressively extracting and enhancing high-frequency sub-bands in different orientations, PHFP substantially improves the model's sensitivity to fine vessel edges and pathological regions (see the red arrows in Figure 11(c)). Meanwhile, the progressive decomposition of low-frequency components enables a more comprehensive representation of the principal vascular structures, thereby reinforcing global contextual connectivity (see the yellow arrows in Figure 11(c)). Moreover, through inverse wavelet transformation, PHFP incrementally re-synthesizes multi-level high- and low-frequency components into the spatial domain, thereby simultaneously preserving intricate structural details and ensuring uninterrupted vascular continuity. In doing so, it effectively alleviates the issues of redundant information and artifact accumulation that are commonly observed in convolution-based feature extraction (see the white arrows in Figure 11(b)).

*3.5. Loss Function Design*

For the ICA vessel segmentation task, where both topological continuity and robustness against noise are critical, we adopt the Mean Squared Error (MSE) as the primary loss function. It is formally defined in Eq. 11.:

$$L_{MSE} = \frac{1}{N}\sum_{i=1}^{N}(y_i - \hat{y}_i)^2, \tag{11}$$

where $N$ denotes the total number of pixels, $y_i$ represents the ground truth label, and $\hat{y}_i$ corresponds to the model prediction.

Importantly, MSE provides a continuous and smooth gradient, which not only facilitates effective suppression of noise interference but also preserves the integrity of vascular topology. During model training, the gradients derived from this loss are propagated backward to optimize network parameters, thereby improving both the accuracy and stability of the segmentation outcomes.

## 4. Experiments

This Section presents the experimental setup, followed by a comprehensive performance analysis encompassing both comparative and ablation studies, as well as the evaluation of stenosis detection. The experimental findings collectively demonstrate the accuracy and robustness of the proposed method.

*4.1. Experimental Setup*

**Training Details**. The proposed SFD-Mamba2Net was implemented using the PyTorch framework and trained on two RTX 3090 GPUs with a total memory of 48 GB. We adopted the Adam optimizer [35] with an initial learning rate of $1 \times 10^{-4}$, momentum parameters $\beta_1 = 0.9$ and $\beta_2 = 0.999$, and a weight decay of $1 \times 10^{-5}$. A hierarchical learning rate scheduling strategy based on StepLR was employed. Specifically, the learning rate was kept constant at $1 \times 10^{-4}$ during the first 10 epochs to ensure comprehensive capture of the global characteristics of major vessels. Between epochs 10 and 30, it was reduced by half every 10 epochs, thereby progressively refining the representation of secondary vascular branches. Beyond 30 epochs, the learning rate was fixed at $3.125 \times 10^{-5}$ to concentrate on precise delineation of microvessel boundaries and preservation of fine-grained structural details, ultimately yielding improved segmentation accuracy and enhanced model generalizability.

**Image Dataset.** The dataset employed in this study was collected from 99 patients who underwent invasive coronary angiography (ICA) at Jiangsu Province People's Hospital, China, between February 26 and July 18, 2019. All images were acquired using the Siemens AXIOM-Artis interventional angiography system at a frame rate of 15 fps, with a spatial resolution of 512 × 512 pixels and an inter-pixel spacing ranging from 0.258 mm to 0.390 mm. The dataset comprises 187 left coronary artery (LCA) images and 127 right coronary artery (RCA) images, each annotated on a frame-by-frame basis by highly experienced interventional cardiologists. For each patient, up to five standard projection angles were selected, and one representative frame per angle was chosen for segmentation labeling. Table 1 provides a detailed overview of the distribution of images across different ICA projection angles.

Table 1: Distribution of ICA views and image counts. (LCA: Left Coronary Artery; RCA: Right Coronary Artery; LAO: Left Anterior Oblique; RAO: Right Anterior Oblique; CRA: Cranial; CAU: Caudal)

| View | LAO+CAU | LAO+CRA | RAO+CAU | RAO+CRA | Total |
|---|---|---|---|---|---|
| LCA | 82 | 82 | 119 | 120 | 403 |
| RCA | 80 | 83 | 23 | 27 | 213 |

**Evaluation Metrics.** To comprehensively evaluate the segmentation performance of the proposed model, eight widely adopted voxel-level metrics were employed: Dice similarity coefficient (Dice), Accuracy (Acc), F1 score (F1), Intersection over Union (IoU), Sensitivity (Sens), Specificity (Spec), 95% Hausdorff distance (HD95), and Average Symmetric Surface Distance (ASSD). These metrics collectively quantify the agreement between predicted segmentation and ground truth by capturing both volumetric overlap and surface distance errors. All metrics were computed independently on a per-slice basis and subsequently averaged over the test set. Their formal mathematical definitions are provided as follows:

$$Dice = \frac{2TP}{2TP+FP+FN}, \tag{12}$$

$$Acc = \frac{TP+TN}{TP+TN+FP+FN}, \tag{13}$$

$$F1 = 2 \cdot \frac{(TP/(TP+FP)) \cdot (TP/(TP+FN))}{TP/(TP+FP)+TP/(TP+FN)}, \tag{14}$$

$$IoU = \frac{TP}{TN+FP+FN}, \tag{15}$$

$$Sens = \frac{TP}{TP+FN}, \tag{16}$$

$$Spec = \frac{TN}{TN+FP}, \tag{17}$$

$$HD95 = max\left\{P_{95}(\min_{b \in B} \| a - b \|), P_{95}(\min_{a \in A} \| b - a \|)\right\}, \tag{18}$$

$$ASSD = \frac{1}{|A|+|B|}\left(\sum_{a \in A} \min_{b \in B} \| a - b \| + \sum_{b \in B} \min_{a \in A} \| b - a \|\right), \tag{19}$$

where $TP$, $FP$, $TN$, and $FN$ denote true positives, false positives, true negatives, and false negatives, respectively. $P_{95}$ represents the 95th percentile of a distance set. The sets $A$ and $B$ correspond to the surface point clouds of the predicted segmentation and the ground truth, respectively, and $\|\cdot\|$ denotes the Euclidean distance.

*4.2. Segmentation Performance Evaluation*

We compared the proposed SFD-Mamba2Net with seven state-of-the-art deep learning segmentation models, including U-Net [6], U-Net++ [36], Swin-Unet [37], DSCNet [38], VM-UNet

[13], Perspective-Unet [20], and CondSeg [21], using the eight aforementioned evaluation metrics to assess and compare their performance.

**Quantitative Evaluation.** Table 2 summarizes the comparative performance of all models across the evaluated metrics. Overall, SFD-Mamba2Net achieves the best performance in Dice (88.10%), F1 (89.22%), IoU (80.81%), Accuracy (98.80%), and Specificity (99.50%), demonstrating its superior segmentation precision, overall classification capability, and effectiveness in suppressing false positives. Notably, DSCNet attains the highest Sensitivity (91.84%), indicating enhanced detection of low-contrast and fine vascular branches, which helps mitigate missed detections. VM-UNet achieves the lowest HD95 value (5.64 pixels), reflecting its robustness in vessel boundary localization and its potential to improve lesion delineation accuracy. Meanwhile, CondSeg performs best in ASSD (1.17 pixels), highlighting its ability to preserve vascular structural continuity and surface smoothness, effectively preventing edge discontinuities and artifacts.

Taken together, SFD-Mamba2Net demonstrates a clear overall superiority over the competing models, achieving both accurate segmentation of major vessels and robust modeling of complex fine-grained structures. Meanwhile, DSCNet, VM-UNet, and CondSeg exhibit their respective strengths in sensitivity, boundary localization, and surface continuity. In contrast, Swin-Unet shows relatively lower performance across all metrics, indicating limited adaptability to coronary artery segmentation tasks. Collectively, these results highlight that SFD-Mamba2Net excels in capturing both global vascular architecture and fine branching details, making it particularly well-suited for high-precision, structurally continuous segmentation of complex coronary artery anatomies.

Table 2: Quantitative comparison of different segmentation models. (↑ means higher is better; ↓ means lower is better)

|  | U-Net | U-Net++ | Swim-Unet | DSCNet | VM-UNet | Perspective-UNet | ConDSeg | SFD-Mamba2Net |
|---|---|---|---|---|---|---|---|---|
| Dice (%) ↑ | 85.03±1.46 | 85.57±1.29 | 78.07±2.11 | 85.06±1.32 | 87.30±1.12 | 86.73±1.25 | 87.02±1.10 | **88.10±1.25** |
| Acc (%) ↑ | 98.47±1.30 | 98.51±1.10 | 96.88±1.21 | 98.65±1.25 | 98.78±1.15 | 98.79±1.22 | 98.73±1.18 | **98.80±1.43** |
| F1 (%) ↑ | 86.11±1.11 | 86.50±0.97 | 72.36±1.23 | 87.51±1.40 | 89.16±1.27 | 88.97±1.32 | 88.61±1.08 | **89.22±1.07** |
| IoU (%) ↑ | 75.97±1.12 | 76.54±0.98 | 63.96±1.44 | 77.41±1.23 | 80.06±1.18 | 80.11±1.09 | 79.80±1.31 | **80.81±1.56** |
| Sens (%) ↑ | 83.49±2.21 | 83.53±2.15 | 73.00±3.02 | **91.84±2.09** | 89.11±1.56 | 85.80±2.03 | 85.46±1.77 | 87.48±1.93 |
| Spec (%) ↑ | 99.38±0.26 | 99.41±0.24 | 98.35±0.41 | 99.12±0.34 | 99.38±0.31 | 99.43±0.27 | 99.48±0.29 | **99.50±0.24** |
| HD95 (pixel) ↓ | 11.62±1.23 | 9.36±1.08 | 21.34±2.18 | 11.91±1.35 | **5.64±0.22** | 7.84±0.31 | 7.93±0.27 | 7.57±0.35 |
| ASSD (pixel) ↓ | 2.06±0.31 | 1.93±0.28 | 3.37±0.42 | 2.01±0.33 | 1.26±0.22 | 1.33±0.34 | **1.17±0.26** | 1.25±0.14 |

**Qualitative Evaluation.** Figures 12 and 13 present the segmentation results and corresponding heatmap comparisons of different models on the left coronary artery (LCA, top two rows) and right coronary artery (RCA, bottom two rows) views. In Figure 12, red regions indicate the predicted vessel areas. Notably, SFD-Mamba2Net demonstrates a clear advantage in preserving complex bifurcations, tortuous arterial paths, and fine microvascular structures (highlighted by yellow arrows). In contrast, other models exhibit deficiencies in regions with small branch discontinuities or background misclassifications (indicated by green arrows and circles). Overall, SFD-Mamba2Net maintains the continuity of vascular anatomy while producing smoother vessel boundaries, underscoring its superior capability in accurately capturing intricate coronary structures.

Compared with Perspective-UNet and VM-UNet, SFD-Mamba2Net exhibits superior vessel structural clarity and boundary accuracy in ICA images, particularly excelling in bifurcation regions and areas of low contrast. As shown in Figures 12 and 13, the attention heatmaps generated by SFD-Mamba2Net form continuous, slender band-like structures that extend along the main vessel trunks and multiple branch levels, almost perfectly overlapping with the ground truth annotations. Notably, in distal ICA branches (highlighted by yellow arrows in Figure 13), SFD-Mamba2Net maintains strong activations (red regions), whereas other models display fragmented or scattered responses in the same regions, indicating limited sensitivity to the overall vascular architecture.

Furthermore, at bifurcation nodes (red arrows in Figure 13), SFD-Mamba2Net produces a clearly defined "Y"-shaped highlighted skeleton, accurately capturing the main vessel trajectory while preserving the continuity of secondary branches, thereby demonstrating precise comprehension of complex topologies. In sharply curved distal segments of the ICA (green arrows in Figure 13), the heatmaps remain continuously highlighted, whereas other models show abrupt interruptions in activation, reflecting insufficient feature capture in high-curvature regions. These qualitative observations are consistent with the superior quantitative metrics reported in Table 2, further confirming SFD-Mamba2Net's outstanding performance in both macroscopic vessel structures and fine-scale branch details.

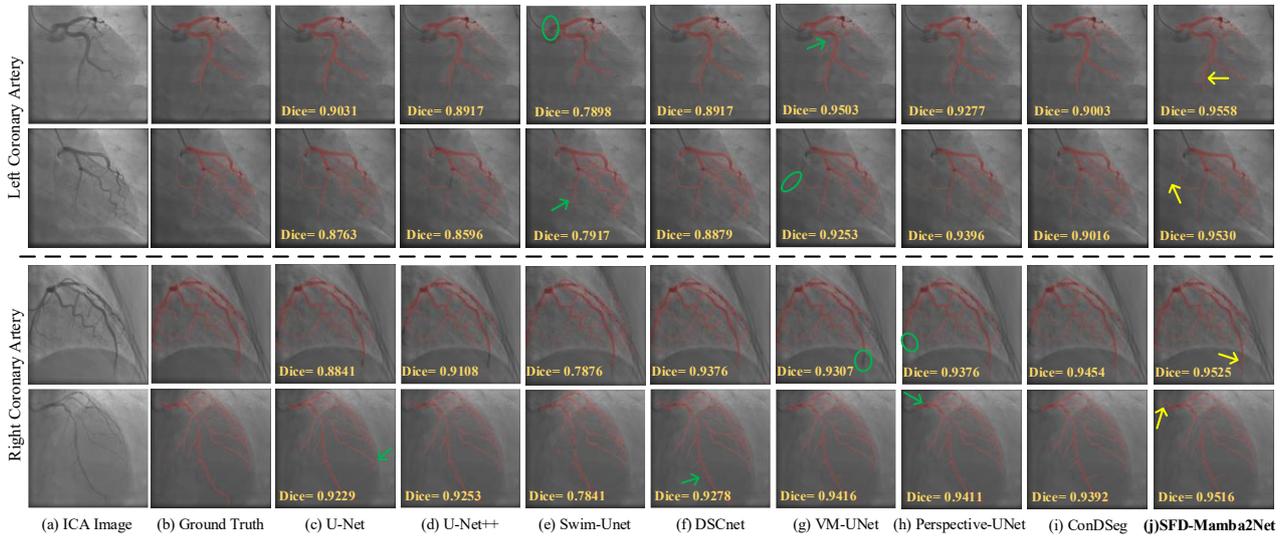

Figure 12. Visual comparison of segmentation results from different models for left (LCA) and right (RCA) coronary arteries. (a) Original ICA images; (b) Ground truth annotations; (c–j) Segmentation outputs of various models, with Dice scores indicated.

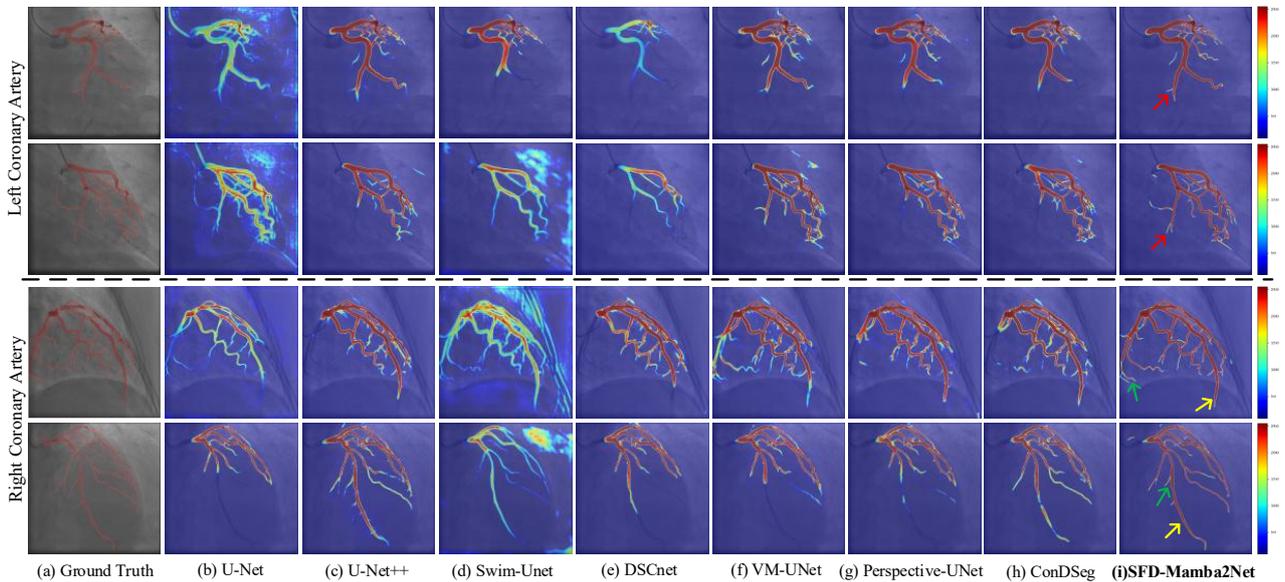

Figure 13. Visual comparison of different models' effectiveness on ICA images. (a) Overlay of the input image and ground truth labels; (b–i) Grad-CAM heatmaps generated by different models, where red indicates high activation and the greatest contribution to the decision, yellow and green indicate moderate activation, and blue denotes minimal

contribution.

*4.3. Ablation Studies*

To systematically evaluate the contributions of individual components within the SFD-Mamba2Net architecture, we conducted ablation experiments across eight model variants: 1) Plain U-Net (U); 2) U-Net augmented with the MASE module only (U+M); 3) U-Net augmented with the AA-DS Mamba2 module only (U+A); 4) U-Net augmented with the PHFP module only (U+P); 5) U-Net incorporating both MASE and AA-DS Mamba2 modules (U+M+A); 6) U-Net incorporating both MASE and PHFP modules (U+M+P); 7) U-Net incorporating both AA-DS Mamba2 and PHFP modules (U+A+P); 8) the full SFD-Mamba2Net integrating MASE, AA-DS Mamba2, and PHFP modules. This systematic design allows for a rigorous assessment of each module's individual and synergistic impact on segmentation performance.

**Quantitative Results.** Table 3 summarizes the performance of the eight aforementioned model variants across multiple evaluation metrics, highlighting the individual contributions of each module to segmentation performance. Introducing the MASE module alone (U+M) yields modest improvements over the baseline U-Net (U) in terms of Dice (85.72% vs. 85.03%) and F1 score, although boundary precision metrics such as HD95 (11.98 vs. 11.62 px) and ASSD (3.07 vs. 2.06 px) show limited improvement. Incorporating the AA-DS Mamba2 module (U+A) further elevates Dice to 86.39% and reduces ASSD to 1.58 px, indicating a substantial enhancement in capturing fine vascular branches and overall structural details. The PHFP module alone (U+P) notably reduces HD95 to 9.78 px, demonstrating improved accuracy in vascular boundary localization.

When these modules are combined (U+M+A, U+M+P, U+A+P), all metrics show further optimization, with HD95 and ASSD decreasing to approximately 1.3-1.5 px, reflecting superior structural continuity and spatial consistency. Ultimately, the full SFD-Mamba2Net, integrating MASE, AA-DS Mamba2, and PHFP modules, achieves the best performance across Dice (88.10%), Acc (98.80%), F1 (89.22%), IoU (80.81%), and Specificity (99.50%), while maintaining HD95 at 7.57 px and ASSD at 1.25 px. These results comprehensively demonstrate our SFD-Mamba2Net's capability to precisely and robustly delineate complex coronary artery topologies in ICA segmentation tasks.

Table 3: Quantitative results of different ablation variants. (↑ means higher is better; ↓ means lower is better)

| | U | U+M | U+A | U+P | U+M+A | U+M+P | U+A+P | SFD-Mamba2Net |
|---|---|---|---|---|---|---|---|---|
| DICE (%) ↑ | 85.03 ± 1.46 | 85.72 ± 1.52 | 86.39 ± 1.27 | 85.65 ± 1.30 | 86.15 ± 1.48 | 86.98 ± 1.41 | 87.34 ± 1.22 | **88.10±1.25** |
| Acc (%) ↑ | 98.47 ± 1.30 | 98.48 ± 1.08 | 98.62 ± 0.91 | 98.52 ± 0.95 | 98.61 ± 1.19 | 98.70 ± 1.22 | 98.75 ± 0.87 | **98.80±1.43** |
| F1 (%) ↑ | 86.11 ± 1.11 | 85.81 ± 1.44 | 87.67 ± 1.42 | 86.82 ± 1.42 | 87.37 ± 1.32 | 88.24 ± 1.27 | 88.82 ± 1.35 | **89.22±1.07** |
| IoU(%) ↑ | 75.97 ± 1.12 | 76.04 ± 1.33 | 78.39 ± 1.16 | 77.01 ± 1.25 | 77.97 ± 1.27 | 79.23 ± 1.34 | 80.20 ± 1.10 | **80.81±1.56** |
| Sens (%) ↑ | 83.49 ± 2.21 | 83.51 ± 1.62 | 86.30 ± 1.34 | 85.55 ± 1.51 | 85.95 ± 1.54 | 86.59 ± 1.66 | 87.51 ± 1.28 | **87.48±1.93** |
| Spec (%) ↑ | 99.38 ± 0.26 | 99.48 ± 0.25 | 99.40 ± 0.26 | 99.32 ± 0.27 | 99.41 ± 0.29 | 99.45 ± 0.28 | 99.48 ± 0.23 | **99.50±0.24** |
| HD95 (pixel) ↓ | 11.62 ± 1.23 | 11.98 ± 2.07 | 8.90 ± 1.57 | 9.78 ± 1.68 | 8.19 ± 1.15 | 8.38 ± 1.12 | **7.10±1.41** | 7.57 ± 0.35 |
| ASSD (pixel) ↓ | 2.06 ± 0.31 | 3.07 ± 0.46 | 1.58 ± 0.28 | 1.77 ± 0.33 | 1.51 ± 0.22 | 1.44 ± 0.20 | 1.31 ± 0.26 | **1.25±0.14** |

**Qualitative Results.** Figure 14 illustrates the segmentation results of the different model variants for both the left and right coronary arteries. Notably, SFD-Mamba2Net (Figure 14(j)) demonstrates the smoothest and most continuous vessel boundaries, particularly at small branches and bifurcations, effectively reducing segmentation artifacts. In contrast, other model variants (Figures 14(c-i)) exhibit varying degrees of vessel discontinuities or misclassified regions (highlighted by green rectangles). Specifically, the MASE module enhances robustness against background noise while preserving the overall vascular structure; the AA-DS Mamba2 module strengthens the detection of fine vessels and detailed edges; and the PHFP module plays a crucial role in maintaining continuity along both the main vessels and minor branches. Importantly, in regions with curved or small branches indicated by red arrows, SFD-Mamba2Net accurately segments vessels that other variants tend to miss or represent ambiguously, highlighting its superior sensitivity to complex vascular topologies.

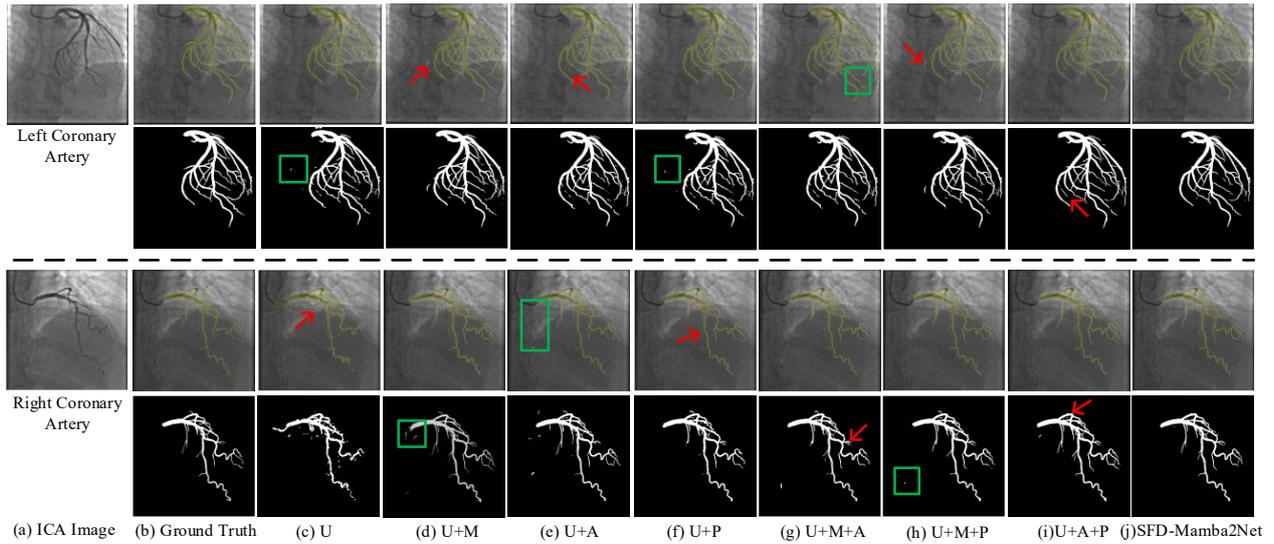

Figure 14. Visualized segmentation results of the left and right coronary arteries. (a) Original ICA images; (b) Corresponding ground truth annotations; (c–i) Segmentation outputs from different ablation models; (j) Segmentation output of SFD-Mamba2Net. Each case consists of two rows: the upper row shows the predicted segmentation (yellow) overlaid on the original image, and the lower row displays the corresponding binary vessel segmentation map.

Figure 15 presents the attention heatmaps generated by each model on coronary angiography images, where red regions indicate highly activated areas that contribute significantly to the model's decisions. Compared with other variants, SFD-Mamba2Net produces attention maps with broader coverage and more precise localization, effectively capturing vessel contours and critical branch pathways. This demonstrates a substantial enhancement in the model's ability to identify semantically important regions.

Integrating the observations from Figures 14 and 15, it is evident that the combination of multi-scale structural priors, state-space duality modeling, and frequency-domain feature enhancement confers clear advantages in maintaining vessel continuity, preserving fine-grained details, and accurately detecting branches. These qualitative findings align closely with the quantitative performance improvements summarized in Table 3.

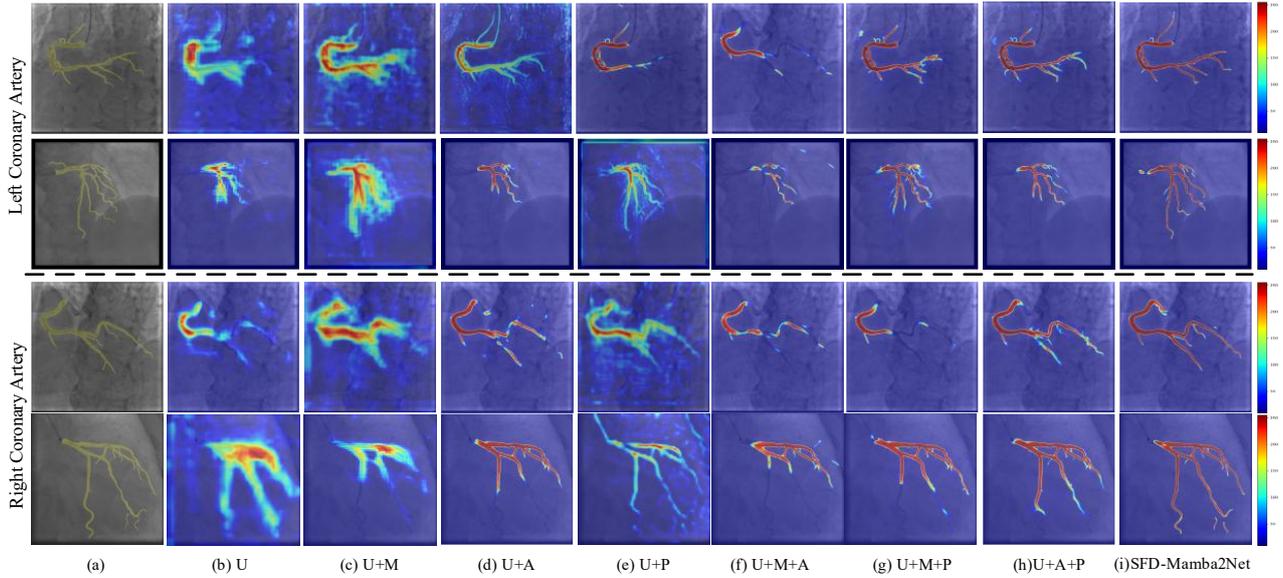

Figure 15. Attention heatmaps of different ablation variants highlighting regions of focus during segmentation. (a) Original ICA images; (b-i) Attention maps from each model variant. Red indicates strong activation, yellow/green indicates moderate activation, and blue indicates low activation regions.

### *4.4. Stenosis Detection and Evaluation*

To evaluate the clinical applicability of the segmentation results, this study introduces a stenosis detection algorithm [39] to quantitatively assess the model's ability to identify vascular narrowing lesions. This pipeline is designed to bridge the gap between anatomical segmentation and clinical diagnostic requirements by determining whether the automatically segmented vessel contours can accurately localize and assess the severity of arterial stenosis, thereby validating the model's potential as a diagnostic aid. Specifically, the algorithm employs a centerline-based analysis to extract the vascular topology, compute lumen diameter variations, and identify regions exhibiting significant constriction. The detailed workflow for coronary artery stenosis detection is described in Algorithm 3.:

| **Algorithm 3:** Artery Stenosis Detection and Evaluation |
| --- |
| **Input:** |
|     Segmented binary artery image $I$ ; |
|     Detection radius $r$; |
|     Centerline length threshold $L_{thresh} = 20$; |
|     Diameter threshold $D_{thresh} = 1.8$ mm ; |
|     Stenosis ratio threshold $b_{thresh} = 0.1$; |
| **Output:** |

| | |
|---|---|
| | Stenosis severity per arterial segment |
| | Evaluation metrics: TPR, PPV, ARMSE, RRMSE |
| Function ExtractCenterline(I): | |
| | Apply morphological thinning on $I$ until skeleton stabilizes |
| | Return single-pixel-width centerline C |
| Function ComputeDiameters(I, C): | |
| | EDT ← EuclideanDistanceTransform($I$) |
| | **For** each point $p \in C$: |
| | $\quad d(p) \leftarrow 2 \times EDT(p)$ |
| | **End** |
| | **Return** diameter map $d$ |
| Function DecomposeSegments(C): | |
| | Label each centerline point by degree: 1 → endpoint, 2 → connector, >2 → bifurcation |
| | Link edges to form segments: $segments \leftarrow EdgeLinking(C)$ |
| | **Return** $segments$ |
| Function DetectStenosis(segments, d): | |
| | **For** each segment S∈ $segments$: |
| | $\quad$ **If** $\max(d(S)) < D_{thresh}$ or $length(S) < L_{thresh}$: continue |
| | $\quad d_{min} \leftarrow \min(\text{local minima of } d(S))$ |
| | $\quad d_{ref} \leftarrow \max(\text{local maxima of } d(S))$ (or $\max(d(S))$ if none) |
| | $\quad b \leftarrow (1 - d_{min} / d_{ref}) \times 100\%$ |
| | $\quad$ **If** $b \geq b\_thresh$: mark $S$ as stenotic with severity $b$ |
| | **End** |
| | **Return** all marked stenotic segments |
| Function MatchWithGroundTruth(stenotic_segments): | |
| | **For** each predicted segment $S_{pred}$: |
| | $\quad$ **If** endpoint distances match any GT segment within radius $r$: mark as matched |
| | $\quad$ **Else** if nearest GT stenotic point within $r$: match accordingly |
| | **End** |
| **Main Routine:** | |
| | C ← ExtractCenterline($I$) |
| | $d \leftarrow$ ComputeDiameters($I, C$) |
| | $segments \leftarrow$ DecomposeSegments($C$) |
| | $stenotic\_segments \leftarrow$ DetectStenosis ($segments, d$) |
| | MatchWithGroundTruth($stenotic\_segments$) |
| | **Return** $stenotic\_segments$ + evaluation metrics |

To quantitatively evaluate the performance of stenosis detection, we employed four commonly used metrics: True Positive Rate (TPR), Positive Predictive Value (PPV), Absolute Root Mean Squared Error (ARMSE), and Relative Root Mean Squared Error (RRMSE), defined as follows:

$$TPR = \frac{TP}{TP+FN}, \tag{20}$$

$$PPV = \frac{TP}{TP+FP}, \tag{21}$$

$$ARMSE = \sqrt{\frac{1}{N}\sum_{n=1}^{N}(b_e - b_g)^2}, \tag{22}$$

$$RRMSE = \sqrt{\frac{1}{N}\sum_{n=1}^{N}(\frac{b_e - b_g}{b_g})^2}, \tag{23}$$

where $b_e$ and $b_g$ denote the model-predicted and ground-truth stenosis rates, respectively, and $N$ represents the number of true positive samples. These metrics collectively provide a comprehensive assessment of the model's accuracy in localizing and quantifying arterial stenosis.

According to the grading criteria of the Society of Cardiovascular Computed Tomography (SCCT) [40], the severity of detected stenoses can be classified into four categories: minimal (1%-24%), mild (25%-49%), moderate (50%-69%), and severe (70%-100%). To enhance the clinical relevance of the evaluation, the following constraints were incorporated in the detection process: segments with a maximum vessel diameter less than 1.8 mm or a centerline length shorter than 20 pixels were considered clinically insignificant; segments lacking inflection points with a second derivative of –2 were regarded as non-stenotic; segments without points having a second derivative of +2 were assigned their global maximum diameter as the reference diameter $d_{ref}$; a vessel segment was classified as exhibiting significant stenosis only if the stenosis rate exceeded 10%. This evaluation framework ensures both the accuracy of stenosis localization and the clinical interpretability of the results, thereby providing a reliable basis for automated detection of coronary artery stenoses.

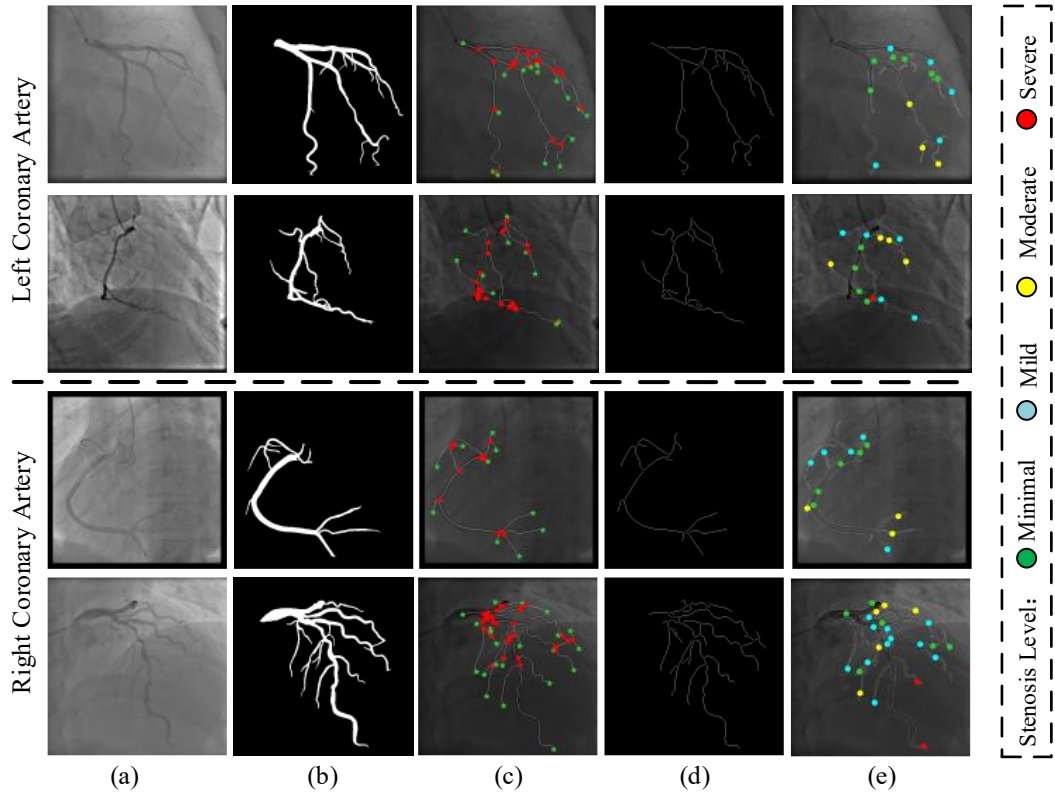

Figure 16. Representative visualization of coronary artery stenosis detection. (a) Original coronary angiography image; (b) Vessel contours predicted by SFD-Mamba2Net; (c) Extracted vessel centerlines; (d) Selected vessel segment centerlines; (e) Detected stenosis points color-coded by severity: green for minimal, blue for mild, yellow for moderate, and red for severe stenosis.

Particularly, the hyperparameter for the detection radius $r$ in Algorithm 3 was empirically set to 10 pixels. Figure 16 illustrates a representative workflow for stenosis detection, encompassing vessel centerline extraction, segment selection, and stenotic point localization, with all vessel contours derived from the predictions of SFD-Mamba2Net. In quantitative evaluation across 132 test images, the algorithm identified a total of 1,792 true positive points, alongside 1,212 false negatives and 1,001 false positives. Table 4 summarizes the performance of stenosis detection based on vessel contours generated by different segmentation models and provides a stratified analysis according to stenosis severity, thereby validating the models' detection capabilities across varying degrees of vascular lesions.

Table 4. Comparison of coronary artery stenosis detection performance across different segmentation models (stratified by stenosis severity).

| Models | Total Stenosis Points | Detected | Detection Ratio | Percent MSE | TPR↑ | PPV↑ | ARMSE↓ | RRMSE↓ |
|---|---|---|---|---|---|---|---|---|
| | | | | | All / Minimal / Mild / Moderate / Severe | | | |
| U-Net | 3993 | 1525 | 0.3819 | 0.1195 | 0.43 / 0.54 / 0.45 / 0.23 / 0.10 | 0.50 / 0.58 / 0.53 / 0.35 / 0.10 | 0.18 / 0.19 / 0.17 / 0.20 / 0.30 | 0.32 / 0.23 / 0.24 / 0.45 / 2.12 |
| U-Net++ | 4062 | 1625 | 0.4093 | 0.1018 | 0.48 / 0.56 / 0.50 / 0.29 / 0.12 | 0.53 / 0.61 / 0.55 / 0.40 / 0.12 | 0.17 / 0.18 / 0.16 / 0.19 / 0.26 | 0.29 / 0.22 / 0.25 / 0.46 / **1.10** |
| Swin-Unet | 3738 | 529 | 0.1415 | 0.1524 | 0.07 / 0.13 / 0.10 / 0.03 / 0.01 | 0.30 / 0.45 / 0.33 / 0.10 / 0.01 | 0.26 / 0.30 / 0.24 / 0.27 / 0.35 | 0.37 / 0.35 / 0.34 / 0.56 / **1.10** |
| DSCNet | 5942 | 1015 | 0.3708 | 0.1169 | 0.20 / 0.22 / 0.22 / 0.18 / 0.15 | 0.14 / 0.10 / 0.17 / **0.57** / 0.22 | 0.23 / 0.16 / 0.18 / 0.33 / 0.65 | 0.53 / 0.17 / 0.27 / 0.79 / 2.50 |
| VM-UNet | 3827 | 1651 | 0.4314 | 0.0972 | 0.56 / 0.57 / 0.53 / 0.39 / 0.29 | 0.63 / 0.59 / 0.64 / 0.47 / 0.35 | 0.16 / 0.15 / **0.12** / 0.19 / 0.36 | 0.41 / 0.18 / 0.20 / 0.44 / **1.53** |
| Perspective-UNet | 4027 | 1406 | 0.3491 | 0.1153 | 0.51 / **0.65** / 0.61 / 0.52 / 0.31 | 0.55 / 0.59 / 0.58 / 0.50 / 0.27 | 0.18 / **0.14** / 0.13 / 0.30 / 0.39 | 0.39 / **0.17** / 0.20 / **0.35** / 1.15 |
| ConDSeg | 4098 | 1478 | 0.3608 | 0.1053 | 0.53 / 0.51 / 0.45 / 0.44 / 0.30 | 0.55 / 0.34 / 0.46 / 0.52 / 0.31 | 0.20 / **0.19** / 0.17 / 0.28 / 0.38 | 0.44 / **0.17** / 0.19 / 0.49 / 2.75 |
| SFD-Mamba2Net | 4005 | 1792 | **0.4474** | **0.08910** | **0.60** / 0.62 / **0.62** / **0.53** / **0.34** | **0.64** / **0.66** / **0.66** / 0.52 / **0.55** | **0.14** / 0.16 / 0.13 / **0.18** / **0.25** | **0.30** / 0.16 / **0.18** / 0.43 / 1.86 |

The results demonstrate that SFD-Mamba2Net consistently outperforms all comparative methods across every evaluation metric. Specifically, it achieves a global ARMSE of 0.14 and RRMSE of 0.30, with TPR and PPV reaching 0.60 and 0.64, respectively, indicating superior accuracy in both stenosis quantification and severity classification. Notably, even within the severe stenosis category, SFD-Mamba2Net maintains relatively high performance (TPR = 0.34, PPV = 0.55), while exhibiting lower ARMSE (0.25) and RRMSE (1.86) compared to most other models.

In contrast, DSCNet and Swin-Unet show marked performance degradation in detecting moderate-to-severe stenoses. For instance, DSCNet records an ARMSE of 0.65 and RRMSE of 2.50 in the severe category, with TPR and PPV dropping to 0.15 and 0.22. Although VM-UNet demonstrates overall competitive performance, its TPR and PPV for severe stenoses remain relatively low (0.29 and 0.35), highlighting limitations in identifying high-risk lesions. Importantly, SFD-Mamba2Net achieves lower ARMSE and RRMSE in moderate-to-severe lesion detection, outperforming Perspective-UNet and ConDSeg and demonstrating enhanced robustness in high-risk stenosis assessment.

These findings indicate that the integration of the MASE module, AA-DS Mamba2 module, and PHFP within SFD-Mamba2Net collectively enhances the network's capability to detect and grade coronary artery stenoses. The improvements are particularly pronounced in complex lesion regions, such as bifurcations, tortuous vessels, and distal microbranches. Overall, the vascular segmentation results produced by SFD-Mamba2Net demonstrate remarkable clinical applicability within stenosis detection pipelines, particularly excelling in the identification and grading of moderate-to-severe vascular stenoses. By simultaneously balancing recall and precision, the proposed framework furnishes robust support for early cardiovascular risk assessment and precision-assisted clinically diagnosis.

## 5. Discussion

### 5.1. Analysis of Artery Segmentation Performance

The proposed SFD-Mamba2Net demonstrates remarkable performance in coronary artery segmentation from invasive coronary angiography (ICA) images. As summarized in Table 2, SFD-Mamba2Net achieves the highest Dice coefficient (88.10%), Accuracy (98.80%), F1 score (89.22%), and IoU (80.81%) among all compared models. Its specificity also reaches a peak of 99.50%, indicating superior stability and reliability in distinguishing vessels from background structures.

With respect to boundary-sensitive metrics, SFD-Mamba2Net attains an HD95 of 7.57 pixels, slightly higher than VM-UNet (5.64 pixels) but notably lower than U-Net++ (9.36 pixels) and ConDSeg (7.93 pixels), reflecting excellent boundary localization. The ASSD is 1.25 pixels, marginally better than VM-UNet (1.26 pixels) and comparable to ConDSeg (1.17 pixels), demonstrating strong geometric consistency and structural continuity. Visualizations of difference maps (Figure 11) further corroborate these findings: false negatives (red) and false positives (blue) are substantially reduced compared to other models, particularly along the main coronary trunks and bifurcations, highlighting SFD-Mamba2Net's superior capacity to preserve boundary continuity and overall vascular structure integrity.

These improvements can be primarily attributed to the synergistic integration of the proposed three key modules. Specifically, the Curvature-Aware Structure Enhancement (CASE) module provides geometric priors at shallow layers, capturing local curvature variations through multi-scale responses. This guides the network to focus on elongated tubular vessels while simultaneously suppressing background noise and enhancing the continuity of low-contrast micro-vessels. The Axial-Alternating Dual-Stream (AA-DS) Mamba2 module, implemented at the bottleneck, employs a dual-path parallel architecture combined with bidirectional State Space Duality (SSD). By efficiently capturing long-range spatial dependencies along both horizontal and vertical paths, it strengthens the global representation of complex vascular topologies and fine branches, effectively mitigating the issue of branch discontinuities. Finally, the Progressive High-Frequency Perception (PHFP) module, positioned in the decoder, progressively enhances high-frequency features via layer-wise wavelet decomposition and deep convolution, while preserving low-frequency global structure. This facilitates multi-scale reconstruction of both main vessels and micro-branches, substantially improving boundary detail recovery and overall structural coherence.

Quantitative metrics further substantiate these gains. For HD95, SFD-Mamba2Net attains 7.57 pixels, outperforming all comparators except VM-UNet (5.64 pixels), thereby indicating lower boundary error. For ASSD, SFD-Mamba2Net achieves 1.25 pixels, performing on par with VM-UNet (1.26 pixels) and competitively close to ConDSeg (1.17 pixels). Taken together, the difference maps and quantitative metrics corroborate that SFD-Mamba2Net delivers high-accuracy segmentation while simultaneously preserving boundary fidelity and vascular structural integrity.

As shown in Figure 17, SFD-Mamba2Net exhibits superior consistency at vascular bifurcations, closely adhering to the ground-truth contours. Considering that clinical stenosis assessment typically prioritizes the coronary trunk rather than capillary-level detail, the structurally continuous vessel delineations produced by SFD-Mamba2Net provide a robust foundation for subsequent diagnostic interpretation and stenosis analysis.

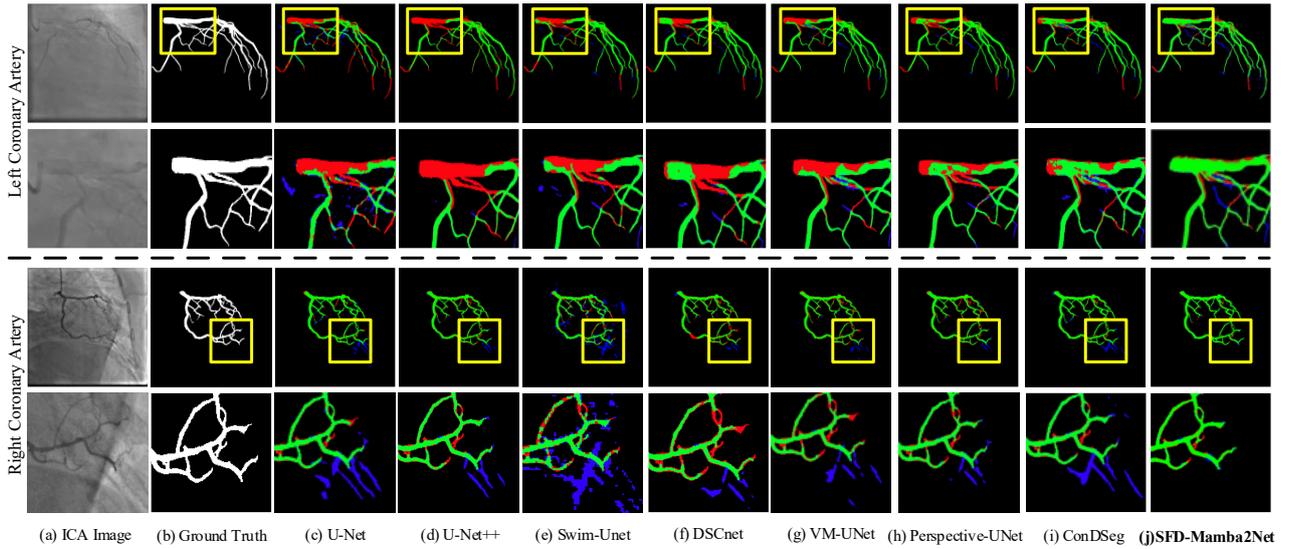

Figure 17. Visualization of segmentation results across different models. (a) Input image; (b) Ground-truth annotation; (c–j) Difference maps of various models. The zoomed-in regions highlight inter-model discrepancies. Green, red, and blue areas denote true positives, false negatives, and false positives, respectively.

*5.2. Analysis of Stenosis Detection Performance*

To evaluate the clinical applicability of SFD-Mamba2Net, we conducted a comprehensive analysis of its segmentation and stenosis-detection performance, integrating qualitative visualizations (Figure 18) with quantitative metrics (Table 4). As illustrated in Figure 18, the model accurately localizes stenotic lesions and stratifies their severity across various coronary locations. A comparison between predicted vessel contours and manual annotations reveals a high degree of topological concordance; notably, the model preserves segmentation continuity and structural integrity in challenging regions such as complex bifurcations and distal micro-vessels. The overlay in Figure 18(e) further underscores this finding: matched stenosis points (white circles with dark-cyan outlines) are densely clustered, whereas false negatives (blue) and false positives (red) are sparse; residual errors are predominantly confined to areas with low vessel contrast or anatomical overlap. Importantly, SFD-Mamba2Net reliably detects subtle, mild, moderate, and severe stenoses, corroborating its potential utility for clinical grading and diagnostic support.

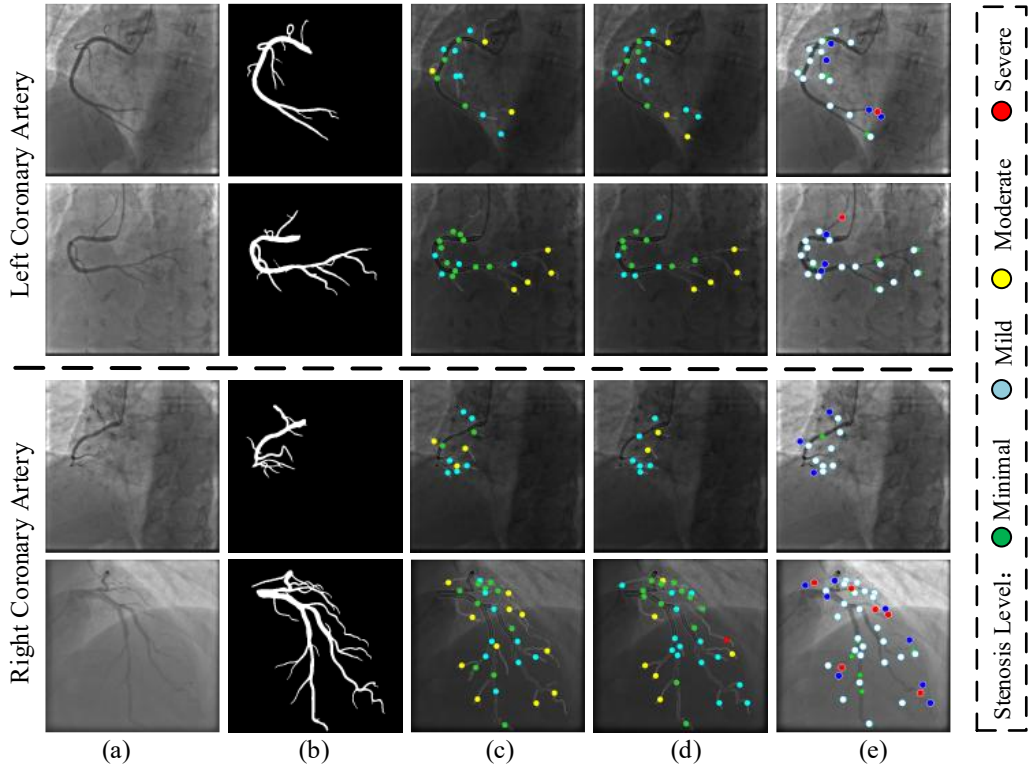

Figure 18. Stenosis detection results. (a) Original image; (b) Ground-truth contours; (c) Detection results based on ground-truth contours; (d) Detection results based on SFD-Mamba2Net contours; (e) Evaluation results. White circles with dark-cyan borders: matched points; blue: missed detections (false negatives); red: false detections (false positives). Green, blue, yellow, and red denote minimal, mild, moderate, and severe stenosis, respectively.

Quantitative results further corroborate the advantages of SFD-Mamba2Net. As reported in Table 4, our model attains the highest overall detection rate (44.74%) and the lowest percentage mean squared error (8.91%) among all compared methods. Stratified analysis by stenosis severity demonstrates that SFD-Mamba2Net consistently maintains elevated true positive rate (TPR) and positive predictive value (PPV) across severity levels; notably, it achieves marked superiority in clinically critical categories of moderate stenosis (TPR = 0.53, PPV = 0.52) and severe stenosis (TPR = 0.34, PPV = 0.55). Specifically, in the severe stenosis, SFD-Mamba2Net outperforms VM-UNet (TPR = 0.29, PPV = 0.35) and Perspective-UNet (TPR = 0.31, PPV = 0.27), indicating enhanced sensitivity and precision for high-risk lesions.

Moreover, the model's absolute RMSE (ARMSE) is tightly controlled within the range 0.14-0.25 across severity levels, while its relative RMSE (RRMSE) remains balanced (0.16-1.86). This demonstrates that SFD-Mamba2Net increases lesion recall without sacrificing error stability.

Collectively, these results indicate that the proposed framework not only detects a larger number of stenotic sites but also provides robust and interpretable quantitative estimates across grading levels, a property that is directly relevant for clinical decision-making, e.g., stent deployment planning or coronary artery bypass grafting strategy.

In summary, SFD-Mamba2Net delivers exceptional performance in graded stenosis detection. Notably, it preserves both a high true-positive rate (TPR) and a high positive-predictive value (PPV) for severe stenoses; moreover, it reliably distinguishes among all severity levels, outperforming competing approaches by a substantial margin. These findings are supported by quantitative metrics and corroborated by qualitative visualizations, underscoring the potential of SFD-Mamba2Net as a practical decision-support tool in clinical workflows.

*5.3. Clinical Overview and Applications*

Coronary artery disease (CAD) remains a major global public-health challenge due to its high rates of morbidity and mortality. ICA is the clinical gold standard for assessing coronary anatomy and stenosis severity, and it plays a central role in treatment planning. However, conventional manual interpretation of ICA is highly operator-dependent and suffers from considerable subjectivity and poor reproducibility [41, 42]. This limitation is exacerbated in the presence of complex lesions (e.g., bifurcations, calcifications, or distal microbranches) and under low-contrast imaging conditions, leading to substantial inter-reader variability in stenosis assessment. Consequently, there is a pressing need for robust, objective and automated ICA analysis tools to improve diagnostic efficiency, reduce observer bias, and better support clinical decision-making.

At the clinical application level, SFD-Mamba2Net demonstrates multiple advantages, of which the vascular segmentation outputs exhibit highly accurate and excellent reproducibility, providing a robust foundation for objective lesion quantification and automated stenosis assessment. Remarkably, the model maintains stable performance even under challenging conditions, such as tortuous vessels, diffuse stenoses, or multi-phase ICA acquisitions, effectively addressing the limitations of conventional manual interpretation. This performance is largely attributable to the model's end-to-end design and the synergistic integration of multiple modules. Specifically, in the shallow layers, the CASE module provides geometric priors and enhances local curvature responses; at the bottleneck, the AA-DS Mamba2 module captures long-range spatial dependencies; and during decoding, the PHFP module progressively reconstructs high-frequency details while integrating low-frequency global structures. Collectively, this hierarchical strategy not only improves segmentation robustness in low-contrast or noisy environments but also enhances edge delineation for microbranches and complex topological regions, thereby ensuring inter-observer consistency and preserving anatomical continuity.

SFD-Mamba2Net can be seamlessly integrated into existing clinical software platforms forCAD diagnosis, providing real-time decision support for diagnosis and interventional procedures. Notably, the model extends stenosis assessment beyond conventional measurements of vessel diameter to a comprehensive analysis of vessel area and morphology, thereby theoretically enhancing its sensitivity to eccentric or irregularly shaped lesions. When combined with high-resolution imaging or three-dimensional reconstruction techniques, SFD-Mamba2Net holds substantial promise for clinical CAD evaluation, risk stratification, and optimization of interventional treatment strategies.

*5.4. Limitations*

Despite the demonstrated excellence of SFD-Mamba2Net in vessel segmentation and stenosis detection, several limitations warrant consideration. First, the scale and diversity of the training dataset remain constrained. Manual annotation of ICA images is both time-consuming and resource-intensive, limiting the availability of large-scale, high-quality labeled data and potentially affecting the model's generalizability across broader clinical settings. Second, the current stenosis detection framework primarily relies on static, single-view ICA images, whereas in clinical practice, multi-view or dynamic image sequences are critical for capturing vessel overlap, foreshortening, and complex spatial relationships. Consequently, overreliance on static views may compromise detection accuracy in such scenarios. Third, ICA inherently represents a two-dimensional projection of three-dimensional coronary arteries; this dimensional simplification unavoidably introduces geometric distortions, which may impact quantitative analysis precision and compromise the reliability of stenosis grading [43].

Future work will focus on addressing these limitations by expanding the dataset to encompass more diverse cases, incorporating multi-view and time-resolved ICA sequences, and exploring three-dimensional reconstruction techniques. These improvements are expected to further enhance the robustness and clinical applicability of SFD-Mamba2Net.

## 6. Conclusion

In this study, we propose SFD-Mamba2Net, an end-to-end deep learning framework designed for ICA images, enabling high-precision coronary artery segmentation and robust stenosis detection. At the shallow layers, the network incorporates a Curvature-Aware Structure Enhancement (CASE) module to provide geometric priors and enhance local curvature features. Within the bottleneck, an Axial-Alternating Dual-Stream Mamba2 (AA-DS Mamba2) module is employed, leveraging dual-path architecture and bidirectional state-space modeling to efficiently capture long-range spatial dependencies, thereby improving the global representation of complex vascular topologies and fine branches. During the decoding stage, a Progressive High-Frequency Perception (PHFP) module is introduced, which progressively enhances high-frequency details through multi-level wavelet decomposition and deep convolution while integrating low-frequency global structures, enabling multi-scale reconstruction of both major coronary trunks and microvascular branches.

Extensive experiments conducted on a large-scale, multi-view ICA image dataset demonstrate that SFD-Mamba2Net significantly outperforms current state-of-the-art models across multiple metrics. Coupled with centerline extraction and stenosis quantification workflows, the network consistently detects mild, moderate, and severe stenoses, providing accurate grading and offering a reliable, objective reference for clinical decision-making.

In conclusion, by synergistically incorporating structural priors, dual-path state-space modeling, and frequency-domain feature enhancement, SFD-Mamba2Net strikes an optimal balance among robustness, accuracy, and interpretability. Beyond serving as a potent instrument for computer-aided diagnosis and therapeutic planning in coronary artery disease, this framework establishes a robust foundation for future investigations into vascular segmentation and stenosis detection within multi-modal and three-dimensional imaging paradigms.

## Acknowledgment

Nan Mu was supported by the Natural Science Foundation of Sichuan Province (2025ZNSFSC1477) and the National Natural Science Foundation of China (62006165).